\documentclass[letterpaper]{article} 
\usepackage[preprint]{aaai2027}  
\usepackage[hyphens]{url}  
\usepackage{graphicx} 
\usepackage{natbib}  
\usepackage{caption} 
\usepackage{subcaption}
\usepackage{algorithm}
\usepackage{algorithmic}

\usepackage{amsmath}
\usepackage{amssymb}

\usepackage{xcolor}
\definecolor{dgreen}{rgb}{0.0,0.55,0.30}
\definecolor{dred}{rgb}{0.80,0.0,0.0}

\usepackage{newfloat}
\usepackage{listings}
\DeclareCaptionStyle{ruled}{labelfont=normalfont,labelsep=colon,strut=off} 
\floatstyle{ruled}
\newfloat{listing}{tb}{lst}{}
\floatname{listing}{Listing}

\usepackage{booktabs}

\title{Beyond the Hard Budget: Sparsity Regularizers for More Interpretable Top-k Sparse Autoencoders}
\author{
    Nathanaël Jacquier\textsuperscript{\rm 1,\rm 2},
    Maria Vakalopoulou\textsuperscript{\rm 4,\rm 5},
    Mahdi S. Hosseini\textsuperscript{\rm 2,\rm 3}
}
\affiliations{
    \textsuperscript{\rm 1}Université Paris-Saclay, CentraleSupélec, France\\
    \textsuperscript{\rm 2}Department of Computer Science and Software Engineering (CSSE), Concordia University, Montreal, QC, Canada\\
    \textsuperscript{\rm 3}Mila--Quebec AI Institute, Montreal, QC, Canada\\
    \textsuperscript{\rm 4}Université Paris-Saclay, CentraleSupélec, Gustave Roussy, INSERM, IHU PRISM, Cancer Data Science Unit, France\\
    \textsuperscript{\rm 5}Université Paris-Saclay, CentraleSupélec, MICS Laboratory, France\\
    nathanael.jacquier@student-cs.fr, mahdi.hosseini@concordia.ca,
    maria.vakalopoulou@centralesupelec.fr
}

\begin{document}

\maketitle

\begin{abstract}
Sparse autoencoders (SAEs) have become a leading tool for interpreting the representations of vision foundation models, decomposing their polysemantic activations into a larger set of sparse, more monosemantic features. The Top-$k$ SAE, a now-standard variant, enforces sparsity architecturally through its activation function, retaining only the $k$ most active latents per input. Because it was designed precisely to avoid the $\ell_1$ penalty used by earlier SAEs and its known drawbacks, it has not been combined with an explicit sparsity regularizer. Yet the Top-$k$ SAE retains limitations of its own, and we hypothesize that a sparsity penalty acting before the selection could sharpen each latent's selectivity and make the code more interpretable, without reintroducing the drawbacks of the $\ell_1$ penalty.We introduce two sparsity regularizers compatible with the Top-$k$ architecture, both acting on the activations before the Top-$k$ selection: an $\ell_1$ penalty on the unselected (off-support) units, and a scale-invariant $\ell_1/\ell_2$-ratio penalty that concentrates the code onto fewer effective units. Both penalties are applied only to the batch-active units, those selected by the Top-$k$ operator at least once within the batch. Across two datasets, three vision foundation models, and a range of $k$, both regularizers consistently improve monosemanticity at no cost to reconstruction quality. The $\ell_1/\ell_2$ penalty further concentrates information into fewer latents, making reconstruction more robust to the inference-time choice of $k$ and improving small-budget linear probing. Our central finding is that hard architectural sparsity and soft sparsity regularization are complementary rather than mutually exclusive.
\end{abstract}

\section{1 Introduction}\label{sec:introduction}
Vision foundation models (VFMs) have become a standard source of general-purpose image embeddings. Yet the embeddings they produce are difficult to interpret, in part because of the \emph{polysemanticity}: a single coordinate of the representation may respond to many seemingly unrelated concepts. A leading explanation for this phenomenon is superposition, whereby a network represents more features than it has dimensions \citep{elhage2022toy}. This opacity is an obstacle to auditing systems built on top of these models.

Sparse autoencoders (SAEs) have recently emerged as a promising tool for interpreting such representations. An SAE encodes an input embedding into a higher-dimensional latent space under a sparsity constraint and reconstructs the input from this sparse code, with the aim of recovering latent units that are as \emph{monosemantic} as possible, ideally, each responding to a single human-interpretable concept. Initially used to interpret the internal activations of large language models (LLMs) \citep{bricken2023monosemanticity, cunningham2023sparse}, SAEs have since been applied to the embeddings of vision foundation models, where they are used to extract and study interpretable visual features \citep{stevens2025sparse, olson2025probing, pach2025sparse}.

SAEs differ in how they impose sparsity. The original formulation, which we refer to as the \emph{vanilla} SAE, adds an explicit $\ell_1$ penalty on the latent code to the reconstruction loss. The $\ell_1$ penalty causes feature shrinkage \citep{gao2024scaling} and produces \emph{dead latents}, units that cease to activate entirely \citep{gao2024scaling}. The Top-$k$ SAE \citep{gao2024scaling}, building on the $k$-sparse autoencoder \citep{makhzani2013ksparse}, was introduced specifically to dispense with the $\ell_1$ penalty: it imposes sparsity architecturally, through an activation function that retains, for each input, only the $k$ latent units with the largest activations and zeroes the rest. This enforces a hard per-sample sparsity budget while avoiding the magnitude shrinkage induced by the $\ell_1$ term.

Yet the Top-$k$ SAE retains its own limitations, which an additional sparsity
term could plausibly address. First, sparsity comes only from selecting the $k$
largest activations, so the objective shapes the code no more than accurate
reconstruction demands, giving the activations no push toward being selective or
concentrated. Second, the value of $k$ is chosen arbitrarily, and a Top-$k$ SAE
tends to overfit to its training $k$: its reconstruction quality degrades when
the number of units retained at inference departs from the training value
\citep{gao2024scaling}.

We introduce two sparsity regularizers that address these limitations, both
acting on the activations \emph{before} the Top-$k$ selection and restricted to
the batch-active units, the units selected by the Top-$k$ operator at least once
within the batch. The first is an $\ell_1$ penalty on the \emph{off-support}
activations, those of units not selected for a given sample, which in a standard
Top-$k$ SAE receive no reconstruction gradient and are left unconstrained.
Penalizing them sharpens each unit's selectivity, encouraging it to activate strongly only where it is among the top contributors and to stay near zero otherwise. We show that this yields more coherent activating image sets and higher monosemanticity. The second penalizes the ratio of the $\ell_1$ to the $\ell_2$
norm of the activations, a scale-invariant sparsity measure introduced by
\citet{hoyer2004nonnegative}. Minimizing it concentrates the code onto fewer
effective units; we show this concentration reduces the model's reliance on the
exact value of $k$ and yields further benefits beyond it.

We evaluate across a range of $k$, using the embeddings produced by three frozen vision foundation models on two datasets, ImageNet-1K \citep{russakovsky2015imagenet} and Open Images V7 \citep{kuznetsova2020open}. The three models are CLIP  ViT-L/14 \citep{radford2021learning}, SigLIP2 \citep{tschannen2025siglip2}, and a supervised ViT-L/16 \citep{dosovitskiy2020image}. We assess interpretability with the Monosemanticity Score of \citet{pach2025sparse}, a label-grounded class-purity measure, and qualitative inspection, and we study the downstream effects of the induced concentration. Our contributions are as follows: (i)~we introduce two sparsity regularizers
compatible with the Top-$k$ architecture, both acting on the pre-selection
activations : an off-support $\ell_1$ penalty and an $\ell_1/\ell_2$-ratio
penalty; (ii)~we show that both regularizers consistently improve
interpretability---measured by monosemanticity and class purity---while
preserving reconstruction quality; and (iii)~we show that the concentration
induced by the $\ell_1/\ell_2$-ratio regularizer addresses a known limitation
of the Top-$k$ SAE, making reconstruction more robust to the
inference-time choice of $k$, and additionally concentrates discriminative
information into fewer leading units, improving small-budget linear probing.


\section{2 Related Work}\label{sec:Related work}

\paragraph{Sparse autoencoders for vision.}
Sparse autoencoders (SAEs) were originally introduced to interpret the internal
activations of large language models \citep{bricken2023monosemanticity}, and are
now also used to extract human-interpretable visual features from the embeddings
of vision foundation models (VFMs) \citep{stevens2025sparse, olson2025probing}.
We follow this line of work and train SAEs on the embeddings of frozen VFMs. A
central goal is to recover latent units that are as \emph{monosemantic} as
possible, each responding to a single human-interpretable concept. To quantify
this for vision SAEs, \citet{pach2025sparse} introduce the Monosemanticity
Score, which measures how similar the images that most strongly activate a given
latent unit are to one another; we adopt it as our primary measure of
interpretability.

\paragraph{Top-$k$ sparse autoencoders.}
SAEs differ in how they impose sparsity. Early SAEs add an explicit $\ell_1$
penalty on the latent code to the reconstruction loss, which causes the
suppression of weaker features \citep{gao2024scaling}. The Top-$k$ sparse
autoencoder \citep{gao2024scaling} instead imposes sparsity architecturally,
through its activation function: for each input it keeps only the $k$ latent
units with the largest activations and sets the remaining ones to zero. We build
directly on this architecture. To the best of our knowledge, no prior work
augments a Top-$k$ sparse autoencoder with an explicit sparsity penalty applied
directly to the activation vector seen by the Top-$k$ operator. This it the gap addressed by the two regularizers we propose.

\section{3 Method}\label{sec:method}

\begin{figure*}[t]
\centering
\includegraphics[width=0.49\textwidth]{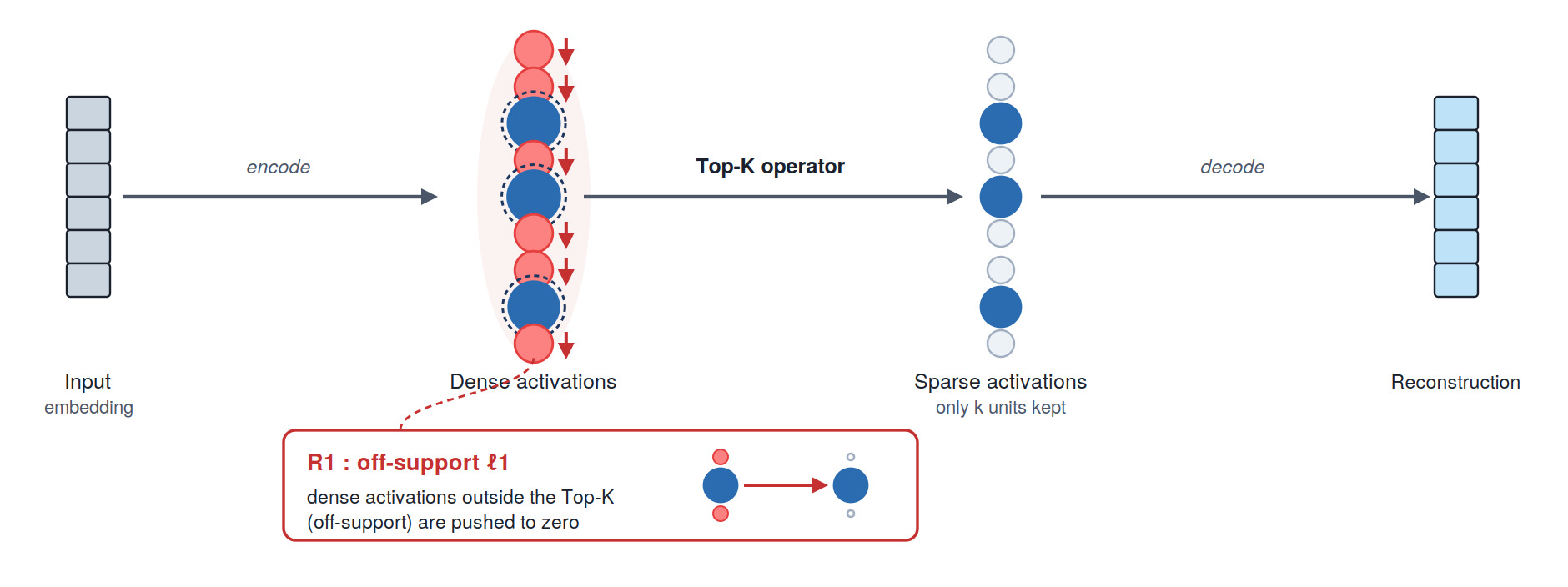}\hfill
\includegraphics[width=0.49\textwidth]{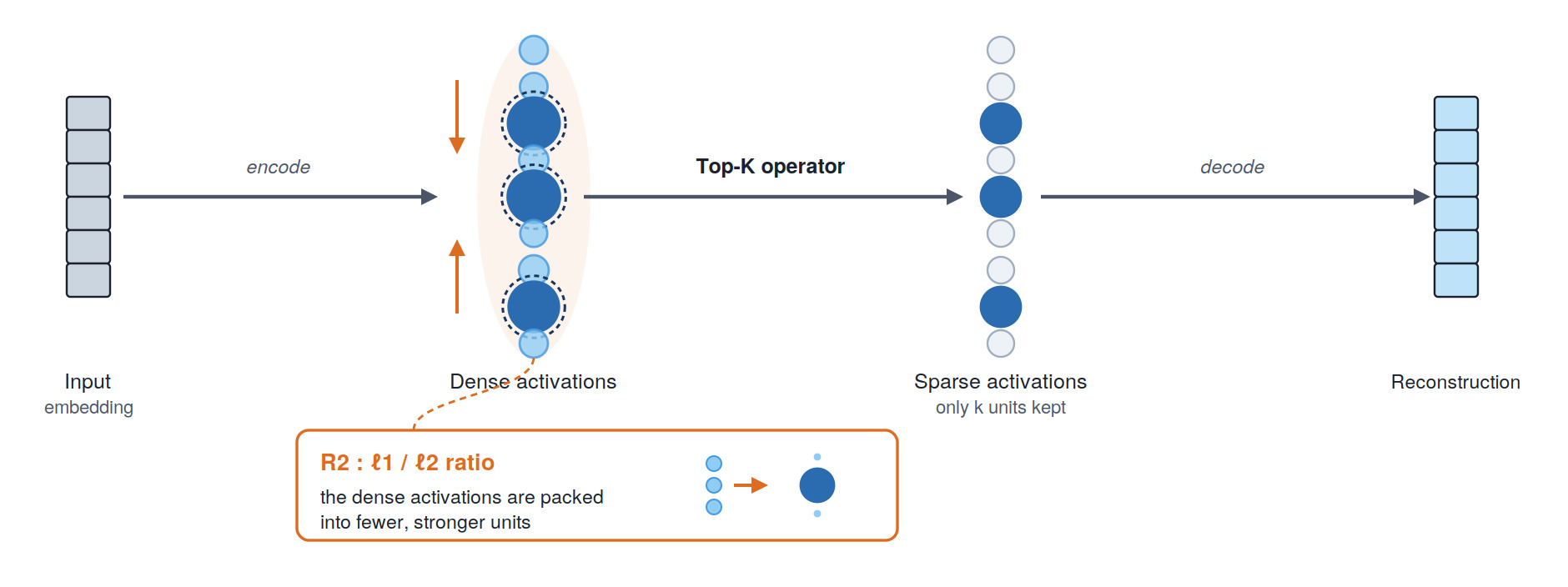}
\caption{Effect of the two regularizers on the pre-selection activations.
\textbf{(left)} Regularizer~1 shrinks the activations not selected by the Top-$k$
operator toward $0$; \textbf{(right)} Regularizer~2 concentrates the activation
vector onto fewer units.}
\label{fig:pipelines}
\end{figure*}

We propose two regularizers compatible with the Top-$k$ architecture, each imposing an additional sparsity constraint on the activations before the Top-$k$ operator. Their effects are illustrated in Figure~\ref{fig:pipelines}.

\subsection{3.1 Notation and Architecture}
We train a Top-$k$ sparse autoencoder on embeddings $x \in \mathbb{R}^d$ produced by a frozen vision foundation model. A batch of $N$ samples is denoted $\mathcal{B} = \{x^{(1)}, \dots, x^{(N)}\}$. The SAE has $m \gg d$ latent units, with encoder weights $W_e \in \mathbb{R}^{m \times d}$, decoder weights $W_d \in \mathbb{R}^{d \times m}$, an encoder bias $b_e \in \mathbb{R}^m$ over the latent space, and a bias $b_d \in \mathbb{R}^d$ over the input space.

For a sample $x^{(i)}$, the encoder first produces the \emph{pre-activations}
\begin{equation}
\pi^{(i)} = W_e\,\big(x^{(i)} - b_d\big) + b_e \in \mathbb{R}^m ,
\end{equation}
to which a ReLU is applied to obtain the \emph{activations}
\begin{equation}
a^{(i)} = \mathrm{ReLU}\!\left(\pi^{(i)}\right) \in \mathbb{R}^m_{\geq 0}.
\end{equation}
The Top-$k$ operator then retains the $k$ largest activations and zeros the rest. Let $S^{(i)} \subseteq \{1, \dots, m\}$, with $|S^{(i)}| = k$, denote the index set of the $k$ largest coordinates of $a^{(i)}$. The \emph{sparse code} is
\begin{equation}
z^{(i)} = \mathrm{TopK}_k\!\left(a^{(i)}\right), \qquad
z^{(i)}_j =
\begin{cases}
a^{(i)}_j, & j \in S^{(i)}, \\
0, & j \notin S^{(i)}.
\end{cases}
\end{equation}
Finally, the decoder reconstructs the input from the sparse code, adding the centering bias back,
\begin{equation}
\hat{x}^{(i)} = W_d\, z^{(i)} + b_d .
\end{equation}

Throughout training we constrain each decoder column to unit $\ell_2$ norm, following standard SAE practice (Bricken et al. 2023; Gao et al. 2024).

\subsection{3.2 Active-Unit Mask}

Both regularizers are restricted to the activations selected by the Top-$k$ operator for at least one sample of the batch. We define this \emph{active set} as
\begin{equation}
\mathcal{A}(\mathcal{B}) = \bigcup_{i=1}^{N} S^{(i)} \subseteq \{1, \dots, m\},
\end{equation}
and encode it as a binary mask $\mu \in \{0,1\}^m$, the per-unit indicator of the active set,
\begin{equation}
\mu_j = \mathbf{1}\big[\, j \in \mathcal{A}(\mathcal{B}) \,\big] =
\begin{cases}
1, & \text{if } \exists\, i \in \{1,\dots,N\}:\ j \in S^{(i)}, \\
0, & \text{otherwise.}
\end{cases}
\end{equation}
The mask is recomputed for each batch and treated as a constant during backpropagation; no gradient flows through $\mu$. Masking restricts the penalty to units that contribute to the reconstruction of the batch. Units that are never selected receive no reconstruction gradient, so penalizing them would drive their activations toward zero with no counteracting signal, inflating the number of dead neurons.

\subsection{3.3 Regularizers}

Both regularizers act on the activations $a^{(i)}$, restricted to the batch-active units by the mask $\mu$. They share the notation above and differ only in their expression. We denote the regularizer evaluated on a batch by $R(\mathcal{B})$. We study the two regularizers separately, each trained and evaluated as an independent alternative to the baseline.

\paragraph{Regularizer 1 (off-support $\ell_1$).} We penalize the $\ell_1$ norm of the masked residual between the activations and the sparse code,
\begin{equation}
R(\mathcal{B}) = \frac{1}{N} \sum_{i=1}^{N} \big\| \mu \odot \big( a^{(i)} - z^{(i)} \big) \big\|_1 ,
\end{equation}
where $\odot$ is the elementwise (Hadamard) product. Since $a^{(i)}_j = z^{(i)}_j$ on the selected units, this residual is supported exactly on the units \emph{not} selected by the Top-$k$ operator (the off-support units); the mask further restricts it to the units active across the batch. Minimizing $R$ thus drives the off-support activation mass of the batch-active units toward zero.

\paragraph{Regularizer 2 ($\ell_1/\ell_2$ ratio).}
We penalize the ratio of the $\ell_1$ to the $\ell_2$ norm of the masked activations,
\begin{equation}
R(\mathcal{B}) = \frac{1}{N} \sum_{i=1}^{N} \frac{\big\| \mu \odot a^{(i)} \big\|_1}{\big\| \mu \odot a^{(i)} \big\|_2 },
\end{equation}
The ratio $\|\cdot\|_1 / \|\cdot\|_2$ is scale-invariant, so the penalty concentrates the code without competing with the reconstruction term over its overall magnitude. Its square is a standard proxy for the effective number of active units \citep{hoyer2004nonnegative}, so minimizing $R$ pushes each code toward fewer effective active units.

\subsection{3.4 Training Objective}

With the reconstruction loss
\begin{equation}
\mathcal{L}_{\text{recon}}(\mathcal{B}) = \frac{1}{N} \sum_{i=1}^{N} \big\| x^{(i)} - \hat{x}^{(i)} \big\|_2^2,
\end{equation}
and the auxiliary Top-$k$ loss $\mathcal{L}_{\text{aux}}(\mathcal{B})$ that reconstructs the residual from the top unselected (dead) units \citep{gao2024scaling}, the total objective is
\begin{equation}
\mathcal{L}(\mathcal{B}) = \mathcal{L}_{\text{recon}}(\mathcal{B}) + \alpha\, \mathcal{L}_{\text{aux}}(\mathcal{B}) + \lambda\, R(\mathcal{B}),
\end{equation}
where $\lambda \geq 0$ controls the strength of the sparsity penalty and $\alpha \geq 0$ the auxiliary term. The baseline Top-$k$ SAE corresponds to $\lambda = 0$.

\begin{table*}[t]
\centering
\small
\setlength{\tabcolsep}{3pt}
\begin{tabular}{lll cccc cccc cccc}
\toprule
& & & \multicolumn{4}{c}{$k=32$} & \multicolumn{4}{c}{$k=64$} & \multicolumn{4}{c}{$k=128$} \\
\cmidrule(lr){4-7} \cmidrule(lr){8-11} \cmidrule(lr){12-15}
Encoder & Data & Method & $R^2$ & M$_\mu$ & M$_m$ & Dead & $R^2$ & M$_\mu$ & M$_m$ & Dead & $R^2$ & M$_\mu$ & M$_m$ & Dead \\
\midrule
CLIP & ImageNet & Baseline & \underline{0.775} & 0.498 & 0.487 & 22 & 0.824 & 0.464 & 0.435 & 4 & 0.873 & 0.424 & 0.382 & 2 \\
 &  & Reg.~1 & 0.774 & \textbf{0.526} & \textbf{0.523} & 22 & \underline{0.824} & \textbf{0.505} & \textbf{0.493} & 1 & \underline{0.874} & \textbf{0.444} & \textbf{0.410} & 1 \\
 &  & Reg.~2 & \textbf{0.777} & \underline{0.503} & \underline{0.495} & 49 & \textbf{0.826} & \underline{0.483} & \underline{0.463} & 12 & \textbf{0.875} & \underline{0.432} & \underline{0.394} & 2 \\
\addlinespace
 & Open Images & Baseline & 0.700 & 0.416 & 0.414 & 18 & 0.758 & 0.386 & 0.373 & 2 & 0.810 & 0.360 & 0.351 & 0 \\
 &  & Reg.~1 & \underline{0.701} & \textbf{0.445} & \textbf{0.451} & 32 & \underline{0.759} & \textbf{0.424} & \textbf{0.418} & 1 & \underline{0.815} & \textbf{0.393} & \textbf{0.380} & 0 \\
 &  & Reg.~2 & \textbf{0.706} & \underline{0.425} & \underline{0.427} & 40 & \textbf{0.767} & \underline{0.401} & \underline{0.391} & 10 & \textbf{0.819} & \underline{0.368} & \underline{0.358} & 2 \\
\midrule
SigLIP2 & ImageNet & Baseline & 0.774 & \underline{0.586} & 0.591 & 177 & 0.827 & \underline{0.576} & 0.575 & 200 & 0.871 & 0.560 & 0.548 & 120 \\
 &  & Reg.~1 & \underline{0.774} & \textbf{0.594} & \textbf{0.600} & 177 & \underline{0.828} & \textbf{0.591} & \textbf{0.591} & 158 & \underline{0.871} & \textbf{0.580} & \textbf{0.570} & 73 \\
 &  & Reg.~2 & \textbf{0.775} & 0.586 & \underline{0.592} & 192 & \textbf{0.830} & 0.575 & \underline{0.576} & 238 & \textbf{0.875} & \underline{0.561} & \underline{0.551} & 159 \\
\addlinespace
 & Open Images & Baseline & 0.685 & 0.536 & 0.554 & 196 & 0.752 & 0.519 & 0.524 & 140 & \underline{0.803} & 0.500 & 0.496 & 69 \\
 &  & Reg.~1 & \underline{0.687} & \textbf{0.546} & \textbf{0.565} & 181 & \underline{0.753} & \textbf{0.545} & \textbf{0.550} & 83 & 0.802 & \textbf{0.528} & \textbf{0.525} & 9 \\
 &  & Reg.~2 & \textbf{0.692} & \underline{0.538} & \underline{0.558} & 260 & \textbf{0.759} & \underline{0.522} & \underline{0.534} & 254 & \textbf{0.808} & \underline{0.503} & \underline{0.500} & 94 \\
\midrule
ViT & ImageNet & Baseline & 0.776 & 0.346 & 0.341 & 0 & 0.808 & 0.239 & 0.220 & 0 & 0.839 & 0.155 & 0.128 & 0 \\
 &  & Reg.~1 & \underline{0.780} & \textbf{0.474} & \textbf{0.476} & 0 & \underline{0.810} & \textbf{0.309} & \textbf{0.294} & 0 & \underline{0.839} & \textbf{0.193} & \textbf{0.169} & 0 \\
 &  & Reg.~2 & \textbf{0.800} & \underline{0.414} & \underline{0.402} & 0 & \textbf{0.824} & \underline{0.289} & \underline{0.263} & 0 & \textbf{0.843} & \underline{0.164} & \underline{0.133} & 0 \\
\addlinespace
 & Open Images & Baseline & \underline{0.659} & 0.189 & 0.171 & 0 & 0.704 & 0.129 & 0.106 & 0 & 0.742 & 0.091 & 0.072 & 0 \\
 &  & Reg.~1 & 0.658 & \textbf{0.290} & \textbf{0.274} & 7 & \underline{0.706} & \textbf{0.217} & \textbf{0.198} & 0 & \underline{0.743} & \textbf{0.151} & \textbf{0.129} & 0 \\
 &  & Reg.~2 & \textbf{0.683} & \underline{0.265} & \underline{0.248} & 9 & \textbf{0.720} & \underline{0.195} & \underline{0.176} & 1 & \textbf{0.750} & \underline{0.120} & \underline{0.100} & 0 \\
\bottomrule
\end{tabular}
\caption{Reconstruction ($R^2$), mean and median monosemanticity (M$_\mu$, M$_m$), and number of dead neurons for the Top-$k$ baseline and our two regularizers---Reg.~1 (off-support $\ell_1$) and Reg.~2 ($\ell_1/\ell_2$ ratio). }
\label{tab:reconstruction_monosemanticity}
\end{table*}

\section{4 Experiments}\label{sec:experiments}
We investigate whether the proposed regularizers improve the interpretability of Top-$k$ SAEs. We train matched pairs of models, with and without a regularizer, and evaluate them quantitatively on (i) reconstruction and monosemanticity and (ii) class purity, followed by a qualitative inspection of neurons across their full activation range.

\paragraph{Setup.}
We train and evaluate Top-$k$ SAEs on image embeddings extracted by three frozen vision foundation models: (i) CLIP ViT-L/14 \citep{radford2021learning}, (ii) SigLIP2 \citep{tschannen2025siglip2}, and (iii) a supervised ViT-L/16 \citep{dosovitskiy2020image}, from the images of two datasets, ImageNet-1K \citep{russakovsky2015imagenet} and Open Images V7 \citep{kuznetsova2020open}. The latent dimension is $8192$, and the only hyperparameter varied across runs is $k \in \{32, 64, 128\}$; all remaining hyperparameters are given in the appendix. For each (encoder, dataset, $k$) configuration, we compare the unregularized baseline ($\lambda = 0$) against each regularized model, holding all other settings identical. All reported results are means over three seeds, with standard deviations reported inline where space permits and in the appendix otherwise. All metrics are computed on the test set, except for label-dependent metrics on ImageNet-1K, which use the validation set, as the test labels are not public.

\paragraph{Reconstruction and monosemanticity.}
The primary goal of an SAE is to expose interpretable features, which we quantify with the Monosemanticity Score. $R^2$ measures reconstruction quality, and monosemanticity is computed per latent unit and reported as its mean (M$_\mu$) and median (M$_m$) across units; for all three, higher is better. We additionally report the number of dead neurons (latent units never activated on the test set), for which lower is better. Results are reported in Table~\ref{tab:reconstruction_monosemanticity}, and within every baseline/Reg.~1/Reg.~2 comparison, the best $R^2$ and monosemanticity values are shown in bold and the second best are underlined.

\begin{figure}[t]
    \centering
    \includegraphics[width=\columnwidth]{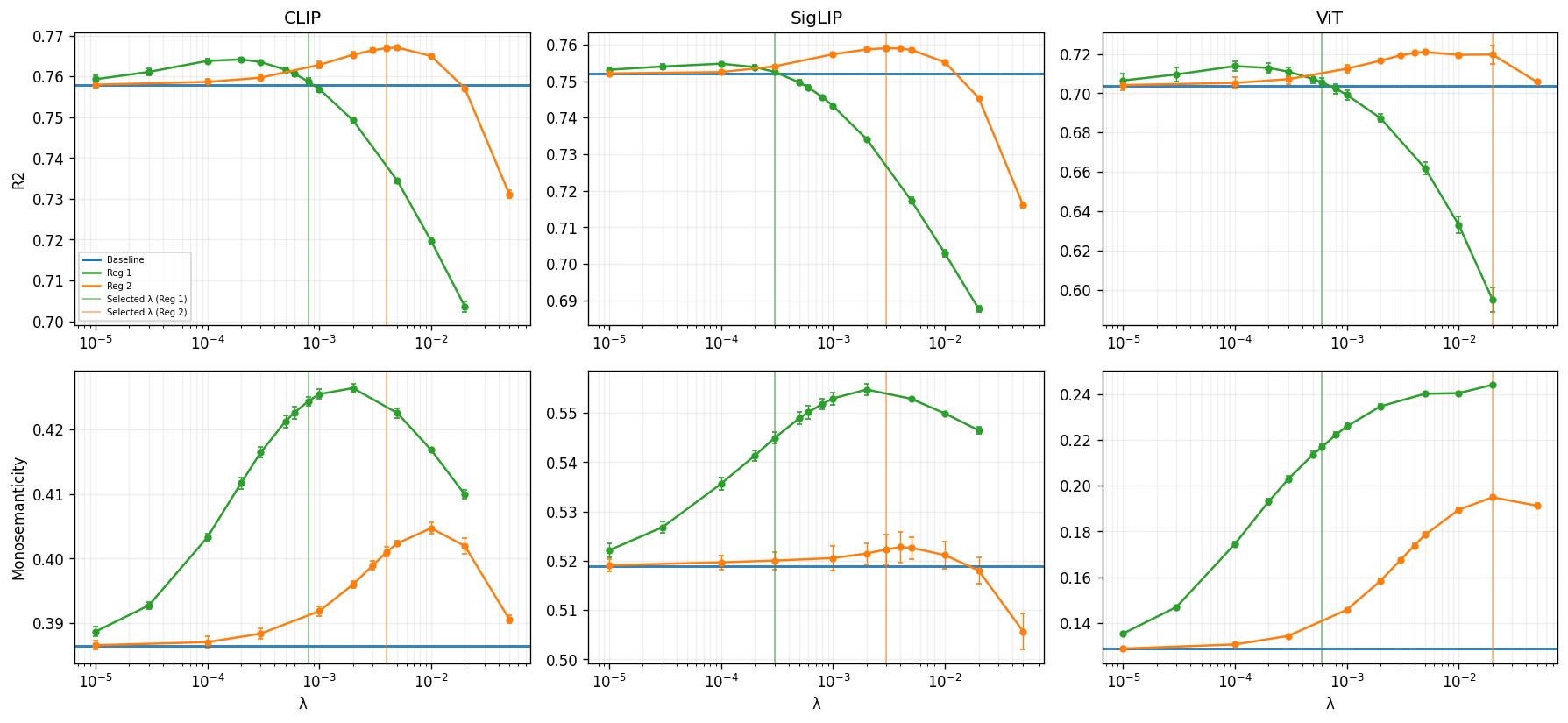}
    \caption{Sensitivity of reconstruction \textbf{(top)} and monosemanticity \textbf{(bottom)} to the
    regularizer strength $\lambda$, on Open Images at $k=64$ for the three
    encoders.}
    \label{fig:lambda_sensitivity}
\end{figure}

For both regularizers, we select the value of $\lambda$ that yields the largest
increase in monosemanticity on the validation set while keeping reconstruction
quality on par with the baseline. Figure~\ref{fig:lambda_sensitivity} shows this
trade-off for a representative configuration (Open Images, $k=64$): both metrics
vary smoothly with $\lambda$ and recover the baseline as $\lambda\!\to\!0$, and
monosemanticity improves over a broad range rather than at an isolated value, so
the reported gains are not sensitive to the precise choice of $\lambda$. The
selected values of $\lambda$ for all configurations are reported in the appendix.

Under this protocol, both regularizers improve monosemanticity while preserving reconstruction quality across all configurations, with two exceptions for SigLIP2 on ImageNet under the Regularizer~2 ($\ell_1/\ell_2$ ratio), where mean monosemanticity decreases marginally.

Finally, the two regularizers have opposite effects on the number of dead neurons: Regularizer~1 tends to decrease it, in some cases substantially (e.g.\ on SigLIP2), whereas Regularizer~2 tends to increase it. This increase is a modest trade-off for the concentration Regularizer~2 induces: pushing mass onto fewer effective units leaves some units unselected more often.

\begin{figure}[t]
    \centering
    \includegraphics[width=\columnwidth]{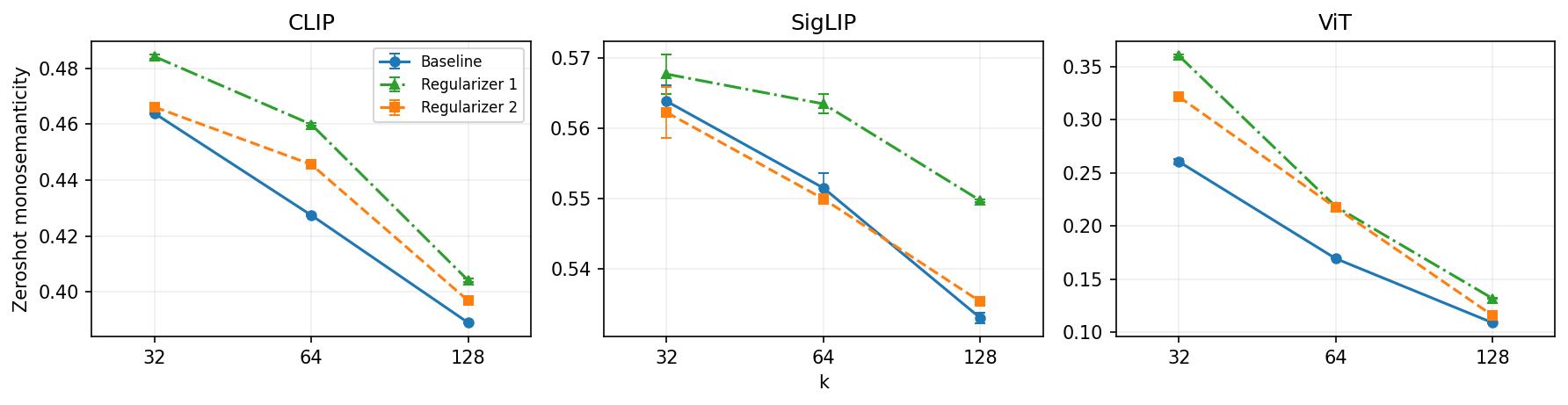}
    \caption{Zero-shot cross-dataset transfer (ImageNet-1K $\to$ Open Images V7):
    mean monosemanticity of SAEs trained on ImageNet-1K and evaluated on Open
    Images V7, versus $k$, for the three encoders.}
    \label{fig:zeroshot_transfer}
\end{figure}

\paragraph{Zero-shot cross-dataset transfer.}
The interpretability gains are not tied to the training distribution. When an
SAE trained on ImageNet-1K is evaluated zero-shot on Open Images V7
(Figure~\ref{fig:zeroshot_transfer}), both regularizers remain more monosemantic
than the baseline in nearly every configuration. Detailed results (Open Images V7 $\to$ ImageNet-1K and reconstructions ($R^2$)) are reported in the appendix. We emphasize that
this shows the gains persist in absolute terms under transfer; we do not claim
the regularizers transfer \emph{better} than the baseline, i.e.\ with a smaller
relative drop.

\begin{table}[t]
\centering
\small
\setlength{\tabcolsep}{3pt}
\begin{tabular}{ll ccc ccc}
\toprule
& & \multicolumn{3}{c}{\textbf{Binary purity}} & \multicolumn{3}{c}{\textbf{Weighted purity}} \\
\cmidrule(lr){3-5} \cmidrule(lr){6-8}
Enc. & $k$ & Base & Reg.~1 & Reg.~2 & Base & Reg.~1 & Reg.~2 \\
\midrule
CLIP & 32 & 0.220 & \textbf{0.278} & \underline{0.232} & 0.290 & \textbf{0.357} & \underline{0.306} \\
 & 64 & 0.150 & \textbf{0.207} & \underline{0.164} & 0.226 & \textbf{0.302} & \underline{0.252} \\
 & 128 & 0.103 & \textbf{0.124} & \underline{0.105} & 0.167 & \textbf{0.201} & \underline{0.176} \\
\midrule
SigLIP2 & 32 & 0.285 & \textbf{0.310} & \underline{0.286} & 0.358 & \textbf{0.382} & \underline{0.358} \\
 & 64 & 0.240 & \textbf{0.267} & \underline{0.242} & 0.331 & \textbf{0.359} & \underline{0.335} \\
 & 128 & \underline{0.188} & \textbf{0.213} & 0.188 & 0.286 & \textbf{0.317} & \underline{0.290} \\
\midrule
ViT & 32 & 0.244 & \textbf{0.384} & \underline{0.310} & 0.417 & \textbf{0.553} & \underline{0.479} \\
 & 64 & 0.127 & \textbf{0.174} & \underline{0.156} & 0.312 & \textbf{0.373} & \underline{0.342} \\
 & 128 & 0.065 & \textbf{0.079} & \underline{0.066} & 0.224 & \textbf{0.257} & \underline{0.228} \\
\bottomrule
\end{tabular}
\caption{Class purity on the ImageNet-1K validation set for the Top-$k$ baseline (Base) and our two regularizers:Reg.~1 (off-support $\ell_1$) and Reg.~2 ($\ell_1/\ell_2$ ratio).}
\label{tab:purity}
\end{table}

\paragraph{Class purity.}
The Monosemanticity Score captures monosemanticity through the similarity of embeddings. To corroborate it with a label-grounded measure, we evaluate the \emph{class purity} of the latent units on the ImageNet-1K validation set. For each active latent $j$, we define its dominant class as the label that fires it most often,
\begin{equation}
c_j^\star = \arg\max_{c} \sum_{n\,:\,y_n = c} \mathbf{1}\!\left[z^{(n)}_j > 0\right],
\end{equation}
where $y_n$ is the label of sample $n$ and $z^{(n)}_j$ its activation on latent $j$. We then summarize how concentrated each latent is on its dominant class with two metrics. \emph{Binary purity} is the fraction of samples activating the latent whose label is $c_j^\star$, counting every activation equally. \emph{Weighted purity} is the same fraction but weighted by activation strength, so that strongly activating samples contribute more; it measures whether a latent's \emph{strongest} responses, in particular, are class-consistent. We average each metric over the active latents and report the mean per configuration.

As shown in Table~\ref{tab:purity}, both regularizers increase binary and weighted purity across all encoders, with a single minor regression in binary purity for SigLIP2 at $k{=}128$ under Regularizer~2. The gains are largest on the ViT-L/16 encoder. Consistent with the monosemanticity results, Regularizer~1 improves purity more than Regularizer~2.

\begin{figure}[t]
\centering
\includegraphics[width=\columnwidth]{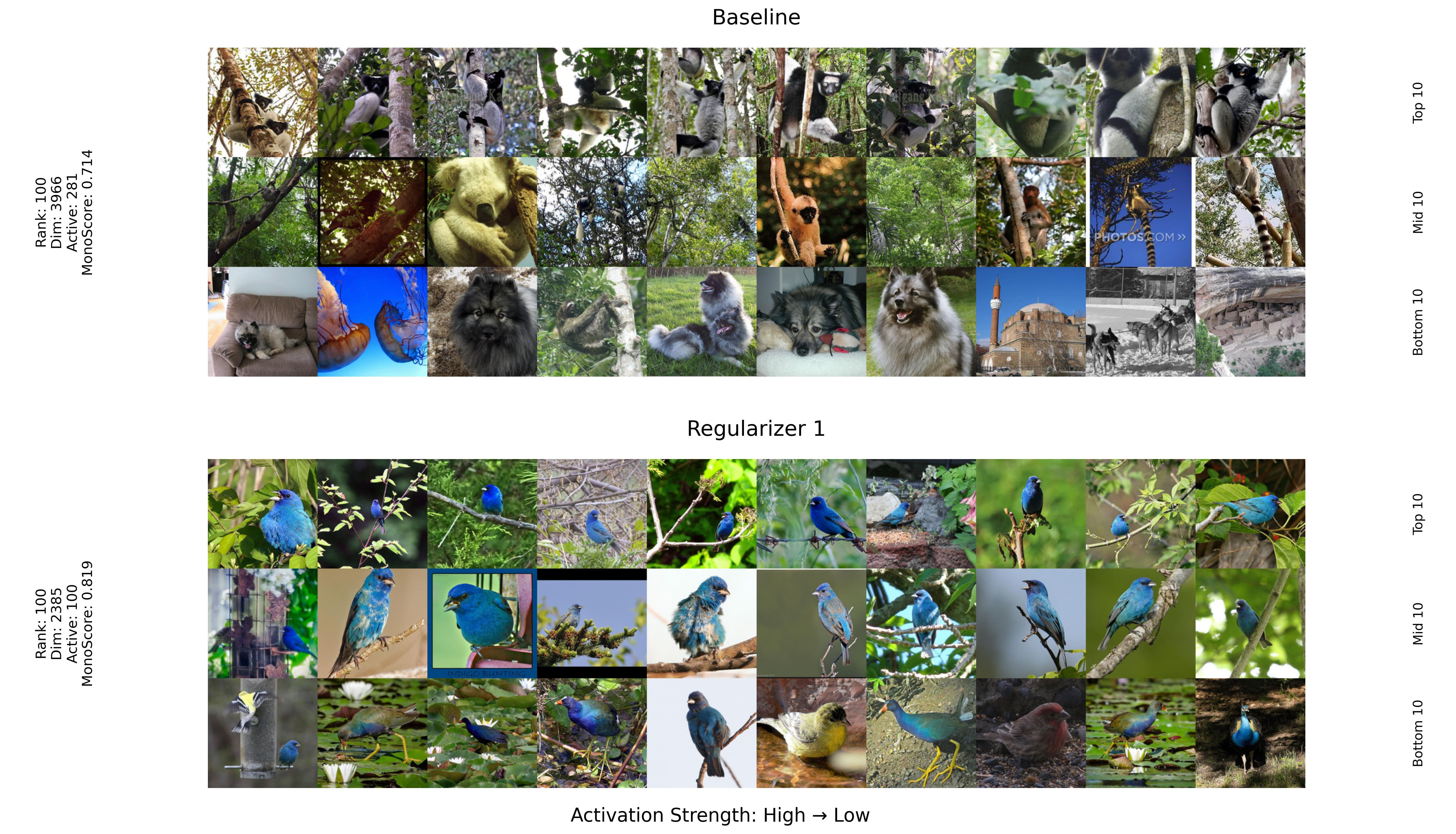}
\caption{Rank-matched qualitative comparison (ViT-L/16, $k=32$). Both blocks show a unit at the same monosemanticity rank (100/8192): top, the baseline (monosemanticity $0.688$); bottom: the Regularizer~1 (monosemanticity $0.805$). In each block, rows show the Top-10, Mid-10, and Bottom-10 activating images. Each row is ordered by decreasing activation strength (high $\rightarrow$ low).}
\label{fig:qual_sameranking}
\end{figure}

\begin{figure}[t]
\centering
\includegraphics[width=\columnwidth]{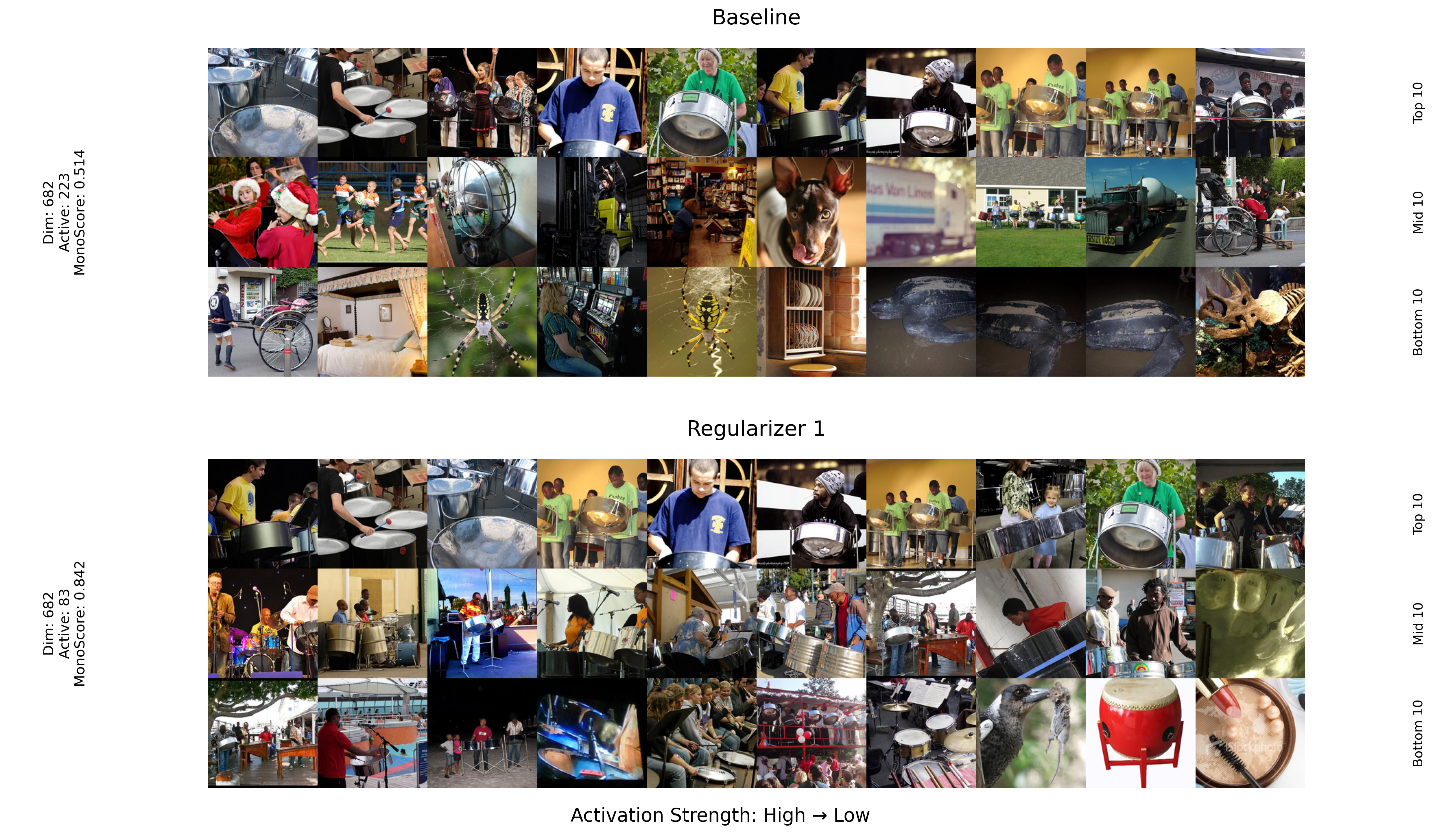}
\caption{Unit-matched qualitative comparison (ViT-L/16, $k=32$). Both blocks show the same unit (unit~4671): top, the baseline (monosemanticity $0.528$); bottom, the same unit trained with Regularizer~1 (monosemanticity $0.865$). In each block, rows show the Top-10, Mid-10, and Bottom-10 activating images.}
\label{fig:qual_samedim}
\end{figure}

\paragraph{Qualitative analysis.}
The Monosemanticity Score is intended to quantify how visually and semantically coherent the images activating a latent are. We complement it with a direct qualitative inspection of the activating images on ImageNet-1K. We focus on the ViT-L/16 encoder with $k=32$ and rank the latents by their test-set monosemanticity score, separately for the baseline and for Regularizer~1 (off-support $\ell_1$). For a given latent, we examine its top-10, middle-10, and bottom-10 activating images, which probes coherence across the full activation range rather than at the peak alone.

We conduct two complementary comparisons, in both cases between two models that differ only in the regularizer and share the same random seed. The first is \emph{rank-matched}: we compare the latent at a given monosemanticity rank with and without the regularizer (in general two distinct units), isolating how coherent a latent of comparable standing is in each model. The second is \emph{unit-matched}: we compare the unit at the same dictionary position with and without the regularizer. Sharing a seed means each position starts from the same initialization, so this tracks a single unit as the penalty reshapes it.

Both comparisons point to the same conclusion. The rank-matched comparison (Figure~\ref{fig:qual_sameranking}) shows that, with the regularizer, the dominant concept of a latent remains consistent down to its weakest (bottom-10) activations, whereas for the baseline coherence degrades away from the top activations. The unit-matched comparison (Figure~\ref{fig:qual_samedim}) shows that a given latent continues to encode the same concept with the regularizer, but its set of activating images becomes markedly more coherent.

\begin{figure}[t]
\centering
\includegraphics[width=\columnwidth]{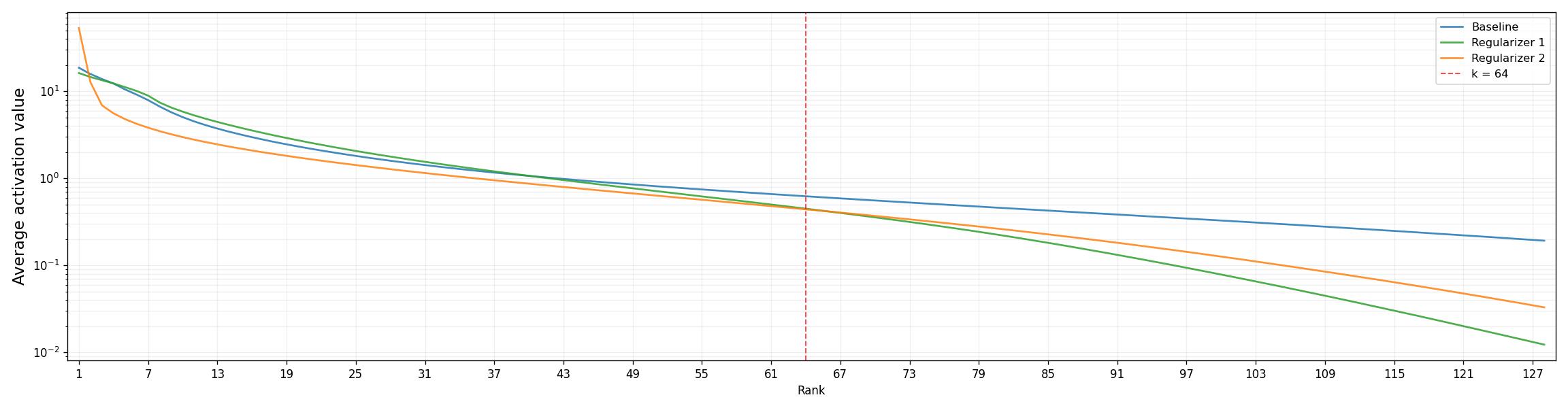}
\caption{Average rank-ordered activation profiles on the ImageNet-1K test set for CLIP at $k=64$. The red line at rank $64$ separates the activations kept by the Top-$k$ operator (left) from those zeroed out (right).}
\label{fig:avg_activations}
\end{figure}

\paragraph{Sparsity effects of the regularizers.}
While the regularizers were shown above to improve interpretability, their original motivation is to impose an additional sparsity constraint. To visualize their effect on the activation distribution, we sort each test sample's $m$ activations in decreasing order and average them rank-by-rank across the ImageNet-1K test set, obtaining the mean magnitude of the $r$-th largest activation as a function of rank $r$. Figure~\ref{fig:avg_activations} plots this averaged, rank-ordered profile for CLIP at $k=64$, for the baseline and for each regularizer.

The two regularizers act on this profile in qualitatively different ways. Regularizer~1 (off-support $\ell_1$) leaves the kept activations almost unchanged relative to the baseline and only suppresses the off-support tail, consistent with its design. Regularizer~2 ($\ell_1/\ell_2$ ratio) instead reshapes the entire profile, concentrating mass onto the very top ranks: its largest activation is substantially higher than the baseline's while the remaining ranks decay much faster. The $\ell_1/\ell_2$ ratio thus has the more pronounced effect on the activation distribution, whereas the off-support penalty acts only where activations are already small.

So far, our analysis showed that Regularizer~1 yields the larger interpretability gains. The activation profiles, however, reveal that Regularizer~2 reshapes the latent geometry more aggressively. The remaining experiments show that Regularizer~2 concentrated geometry brings substantial benefits beyond monosemanticity.

\begin{figure*}[t]
\centering
\includegraphics[height=4.3cm]{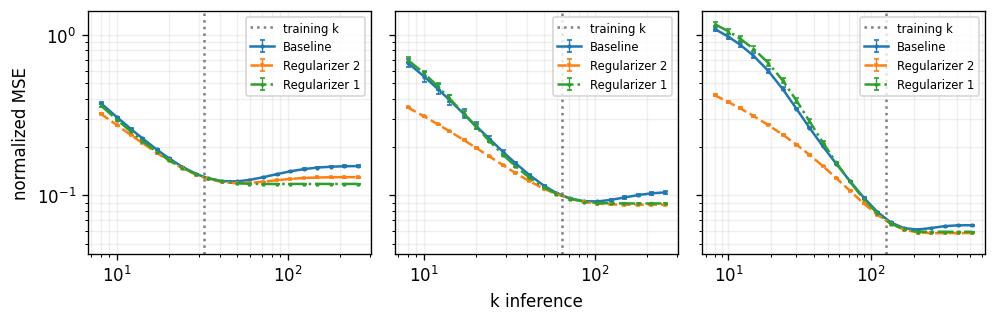}\hfill
\includegraphics[height=4.3cm]{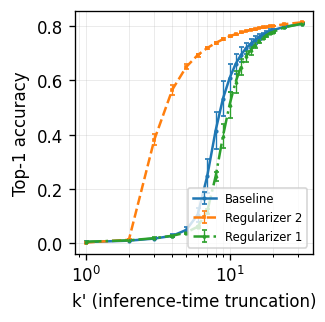}
\caption{Two consequences of the concentration induced by Regularizer~2 ($\ell_1/\ell_2$ ratio), on ImageNet-1K with CLIP; baseline and Regularizer~1 (off-support $\ell_1$) shown for comparison. \textbf{(left)} Robustness to the inference-time number of retained units: each panel is a model trained at a fixed $k$ (left to right: $k=32, 64, 128$; dotted line) and evaluated while varying $k_{\text{inf}}$ at inference; both axes are logarithmic. \textbf{(right)} Probing under activation truncation at $k=64$: a linear probe is trained on codes truncated to their $k'$ largest activations, and top-1 validation accuracy is plotted against $k'$ (log scale).}
\label{fig:reg2_robustness_probing}
\end{figure*}

\paragraph{Robustness to the inference-time $k$.}
Because Regularizer~2 concentrates most of the reconstructed signal onto a few units (Figure~\ref{fig:avg_activations}), the exact number of units retained at inference should matter less than for the baseline, mitigating the known sensitivity of Top-$k$ SAEs to the arbitrary choice of $k$ and their overfitting to the training value \citep{gao2024scaling}. To test this, we train Top-$k$ SAEs on ImageNet-1K at $k \in \{32, 64, 128\}$, for the baseline and both regularizers, and report the normalized reconstruction error (NMSE) as the number of retained units $k_{\text{inf}}$ is varied around the training value.

Figure~\ref{fig:reg2_robustness_probing} reveals two regimes, both more
pronounced at larger training $k$; it shows CLIP, and the corresponding
results for SigLIP2 and the supervised ViT-L/16 are given in
Appendix. Under truncation ($k_{\text{inf}} < k$), only Regularizer~2 improves over the baseline, attaining a consistently lower NMSE with a gap that widens as truncation becomes more aggressive; Regularizer~1 tracks the baseline here, as its off-support penalty leaves the kept activations almost unchanged. Above the training value ($k_{\text{inf}} > k$), the baseline NMSE rises back up, revealing its overfitting to exactly $k$ units, whereas \emph{both} regularizers stay essentially flat. The concentrated geometry of Regularizer~2 is thus the more robust overall, but both penalties remove the baseline's sensitivity to over-retention.

\paragraph{Probing under activation truncation.}
If Regularizer~2 concentrates activation magnitude onto fewer units, it should also concentrate the \emph{information} carried by the code, so that fewer units suffice to identify the class of a sample. We test this by probing the code under activation truncation. For each trained SAE, we form a truncated code by keeping only the $k'$ largest activations of each sample and zeroing the rest, then fit a linear probe on these codes for ImageNet-1K classification ($1000$ classes), training on the training split and reporting top-1 accuracy on the validation split. Sweeping $k'$ yields an accuracy-versus-$k'$ curve; a steep rise at small $k'$ indicates that discriminative information is packed into few coordinates. We probe the baseline and both regularizers, with Regularizer~1 serving as a control: since it does not reshape the retained activations much, it should not concentrate information.

Figure~\ref{fig:reg2_robustness_probing} shows the curves for CLIP at
$k=64$. All three models reach the same asymptotic accuracy (${\approx}0.82$),
so none changes the total discriminative information recoverable from the code;
they differ only in how few units are needed to reach it. Regularizer~1 tracks
the baseline across the sweep, confirming that suppressing the off-support tail
alone does not front-load discriminative information into the leading units.
Regularizer~2, by contrast, reaches high accuracy with far fewer units. Since concentration
is the property under study here, we restrict
the following quantitative analysis to Regularizer~2.

To quantify the concentration gains of Regularizer~2, we report the normalized area under the accuracy-versus-$k'$ curve, defined as the AUC over $k' \in [1, K]$ divided by $K$, which places it in $[0,1]$. A higher value indicates that accuracy is attained at smaller truncation levels. We set $K = 32$, since beyond this point the curves have almost converged and their difference adds no discriminative signal. Table~\ref{tab:trunc_probing} reports this AUC across all nine configurations, together with the top-1 accuracy at the full budget $k'=32$. The AUC is higher for Regularizer~2 in every configuration, confirming that it consistently front-loads discriminative information into the leading activations.

\begin{table}[t]
\centering
\small
\setlength{\tabcolsep}{3pt}
\begin{tabular}{ll cc cc}
\toprule
& & \multicolumn{2}{c}{Norm.\ AUC} & \multicolumn{2}{c}{Top-1 @ $k'{=}32$} \\
\cmidrule(lr){3-4} \cmidrule(lr){5-6}
Enc. & $k$ & Base & Reg.~2 & Base & Reg.~2 \\
\midrule
CLIP & 32 & 0.685\,\tiny{$\pm$\,0.006} & \textbf{0.731\,\tiny{$\pm$\,0.000}} & 0.815\,\tiny{$\pm$\,0.000} & \textbf{0.815\,\tiny{$\pm$\,0.001}} \\
 & 64 & 0.601\,\tiny{$\pm$\,0.014} & \textbf{0.725\,\tiny{$\pm$\,0.002}} & 0.809\,\tiny{$\pm$\,0.001} & \textbf{0.815\,\tiny{$\pm$\,0.001}} \\
 & 128 & 0.488\,\tiny{$\pm$\,0.008} & \textbf{0.693\,\tiny{$\pm$\,0.001}} & 0.775\,\tiny{$\pm$\,0.002} & \textbf{0.805\,\tiny{$\pm$\,0.001}} \\
\midrule
SigLIP2 & 32 & 0.717\,\tiny{$\pm$\,0.019} & \textbf{0.749\,\tiny{$\pm$\,0.036}} & 0.856\,\tiny{$\pm$\,0.001} & \textbf{0.856\,\tiny{$\pm$\,0.001}} \\
 & 64 & 0.627\,\tiny{$\pm$\,0.008} & \textbf{0.759\,\tiny{$\pm$\,0.014}} & 0.853\,\tiny{$\pm$\,0.001} & \textbf{0.856\,\tiny{$\pm$\,0.001}} \\
 & 128 & 0.499\,\tiny{$\pm$\,0.014} & \textbf{0.756\,\tiny{$\pm$\,0.002}} & 0.810\,\tiny{$\pm$\,0.006} & \textbf{0.852\,\tiny{$\pm$\,0.000}} \\
\midrule
ViT & 32 & 0.759\,\tiny{$\pm$\,0.002} & \textbf{0.783\,\tiny{$\pm$\,0.000}} & 0.801\,\tiny{$\pm$\,0.001} & \textbf{0.804\,\tiny{$\pm$\,0.000}} \\
 & 64 & 0.751\,\tiny{$\pm$\,0.007} & \textbf{0.782\,\tiny{$\pm$\,0.000}} & 0.801\,\tiny{$\pm$\,0.001} & \textbf{0.801\,\tiny{$\pm$\,0.001}} \\
 & 128 & 0.738\,\tiny{$\pm$\,0.013} & \textbf{0.754\,\tiny{$\pm$\,0.001}} & 0.801\,\tiny{$\pm$\,0.001} & \textbf{0.801\,\tiny{$\pm$\,0.001}} \\
\bottomrule
\end{tabular}
\caption{Probing under activation truncation on ImageNet-1K, for the Top-$k$ baseline and Regularizer~2 ($\ell_1/\ell_2$ ratio).}
\label{tab:trunc_probing}
\end{table}

\paragraph{Necessity of the active-unit mask.}
Both regularizers are restricted to the batch-active units. The
motivation is that an unselected unit receives no gradient from the
reconstruction term, so penalizing it applies a one-sided force that eventually kills the unit. To test this, we repeat every
configuration of Table~\ref{tab:reconstruction_monosemanticity} with the penalty
applied to all units (no mask). Table~\ref{tab:mask_ablation} confirms the claim:
removing the mask increases the number of dead neurons in nearly every
configuration.

\begin{table}[t]
\centering
\small
\setlength{\tabcolsep}{3pt}
\begin{tabular}{lll cc cc}
\toprule
& & & \multicolumn{2}{c}{ImageNet-1K} & \multicolumn{2}{c}{Open Images V7} \\
\cmidrule(lr){4-5} \cmidrule(lr){6-7}
Enc. & $k$ & Method & Masked & No mask & Masked & No mask \\
\midrule
CLIP   & 32  & Reg.~1 & 25  & 933  & 40  & 2514 \\
       &     & Reg.~2 & 57  & 275  & 45  & 801  \\
       & 64  & Reg.~1 & 1   & 269  & 2   & 837  \\
       &     & Reg.~2 & 10  & 255  & 11  & 319  \\
       & 128 & Reg.~1 & 0   & 24   & 0   & 423  \\
       &     & Reg.~2 & 1   & 65   & 5   & 150  \\
\midrule
SigLIP2 & 32  & Reg.~1 & 202 & 947  & 182 & 1438 \\
       &     & Reg.~2 & 219 & 380  & 224 & 1107 \\
       & 64  & Reg.~1 & 170 & 563  & 77  & 1205 \\
       &     & Reg.~2 & 240 & 375  & 257 & 858  \\
       & 128 & Reg.~1 & 72  & 276  & 7   & 600  \\
       &     & Reg.~2 & 154 & 244  & 84  & 191  \\
\midrule
ViT    & 32  & Reg.~1 & 0   & 615  & 12  & 2426 \\
       &     & Reg.~2 & 0   & 490  & 9   & 2052 \\
       & 64  & Reg.~1 & 0   & 0    & 0   & 391  \\
       &     & Reg.~2 & 0   & 0    & 0   & 617  \\
       & 128 & Reg.~1 & 0   & 0    & 0   & 12   \\
       &     & Reg.~2 & 0   & 0    & 0   & 12   \\
\bottomrule
\end{tabular}
\caption{Number of dead neurons with and without the active-unit mask, for Regularizer~1 (off-support $\ell_1$) and Regularizer~2 ($\ell_1/\ell_2$ ratio). Unlike the other experiments, this ablation is run on a single seed (42), as the effect spans one to two orders of magnitude and is unambiguous across all configurations.}
\label{tab:mask_ablation}
\end{table}

\begin{figure}[t]
    \centering
    \includegraphics[width=\columnwidth]{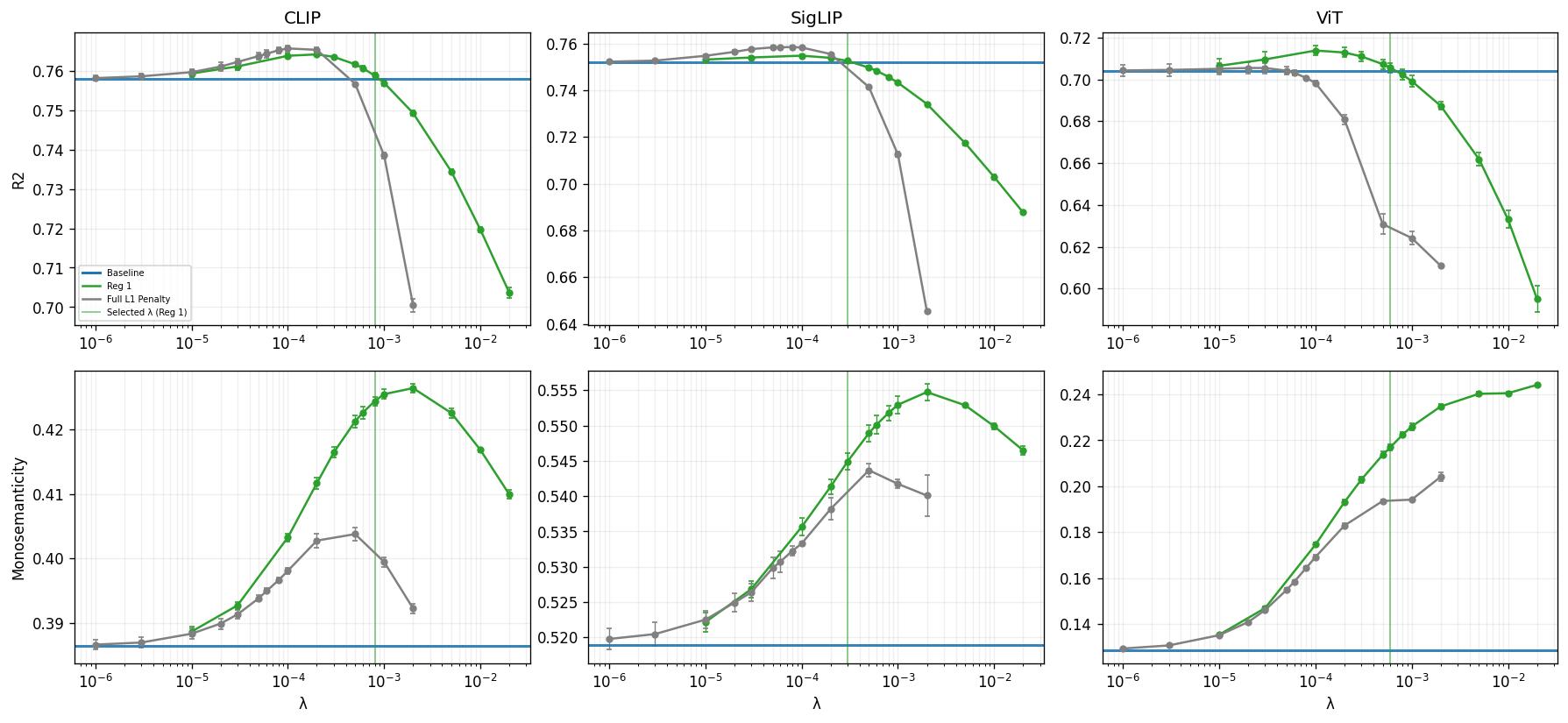}
    \caption{Off-support $\ell_1$ (Reg.~1) versus a full $\ell_1$ penalty, on
    Open Images at $k=64$. Top: $R^2$; bottom: monosemanticity.}
    \label{fig:ablation_full_l1}
\end{figure}

\paragraph{Off-support restriction versus a full $\ell_1$ penalty.}
Regularizer~1 penalizes only the off-support activations. To isolate the effect
of this restriction, we compare it against a full $\ell_1$ penalty on all
activations, kept restricted to the batch-active units so that the two differ
only in whether the selected units are penalized. Sweeping $\lambda$ on Open
Images at $k=64$ (Figure~\ref{fig:ablation_full_l1}), the full penalty degrades
$R^2$ far more sharply as $\lambda$ grows, since it also suppresses the selected
units that carry the reconstruction, and its monosemanticity peaks lower.

\section{5 Conclusion}\label{sec:conclusion}
We introduced two sparsity regularizers compatible with the Top-$k$ architecture: an $\ell_1$ penalty on the off-support units and an $\ell_1/\ell_2$-ratio penalty on the full activation vector. Across two datasets, three vision foundation models, and a range of $k$, both regularizers consistently improve interpretability at no cost to reconstruction quality. Beyond interpretability, we showed that the concentration induced by the $\ell_1/\ell_2$-ratio penalty makes reconstruction more robust to the inference-time choice of $k$ and front-loads discriminative information into the leading units, improving small-budget linear probing.

These results carry a simple message: although the Top-$k$ SAE was introduced
specifically to avoid the $\ell_1$ penalty and its drawbacks, it still benefits
from an explicit sparsity constraint, provided the constraint acts on the
pre-selection activations rather than the sparse code and is restricted to the
batch-active units. Hard architectural sparsity and soft regularization are thus
complementary, not mutually exclusive.

Several directions follow naturally. The regularizers could be extended to other members of the Top-$k$ family, such as BatchTopK \citep{bussmann2024batchtopk} and Matryoshka SAEs \citep{bussmann2025matryoshka}. More broadly, the $\ell_1/\ell_2$ ratio remains underused in representation learning, and our results suggest it is a promising and largely unexplored tool for shaping the geometry of sparse codes.

\bibliography{aaai2027}

\clearpage
\appendix
\renewcommand{\thesection}{A\arabic{section}}
\renewcommand{\thetable}{A\arabic{table}}
\renewcommand{\thefigure}{A\arabic{figure}}
\setcounter{section}{0}
\setcounter{table}{0}
\setcounter{figure}{0}

\begin{center}
    {\Large\bfseries Technical Appendix\par}
\end{center}
\vspace{1em}

\section{Metric Definitions}
\label{app:metrics}

\label{app:metrics}

\paragraph{Notation.}
Let $\mathcal{D} = \{(x^{(i)}, y^{(i)})\}_{i=1}^{N_{\mathcal{D}}}$ denote the
evaluation set, where $x^{(i)} \in \mathbb{R}^d$ is the input embedding and
$y^{(i)} \in \{1,\dots,C\}$ its class label; $N_{\mathcal{D}} = |\mathcal{D}|$
is the number of evaluation samples. Given a trained autoencoder, we obtain for each sample the sparse
code $z^{(i)} \in \mathbb{R}^m$ and reconstruction $\hat{x}^{(i)} \in
\mathbb{R}^d$, as defined in Section~3.1. For latent $j$, its \emph{activating
set} is
\[
A_j = \{\, i \in \{1,\dots,N_{\mathcal{D}}\} : z^{(i)}_j > 0 \,\},
\]
and the set of active latents is
$\mathcal{J}_{\mathrm{act}} = \{\, j \in \{1,\dots,m\} : |A_j| > 0 \,\}$.

\paragraph{Reconstruction ($R^2$).}
We measure reconstruction fidelity with the coefficient of determination,
\[
R^2 = 1 - \frac{\sum_{i=1}^{N_{\mathcal{D}}} \lVert x^{(i)} - \hat{x}^{(i)} \rVert_2^2}
{\sum_{i=1}^{N_{\mathcal{D}}} \lVert x^{(i)} - \bar{x} \rVert_2^2},
\qquad
\bar{x} = \frac{1}{N_{\mathcal{D}}}\sum_{i=1}^{N_{\mathcal{D}}} x^{(i)} .
\]
Higher $R^2$ indicates better reconstruction quality.

\paragraph{Monosemanticity score.}
We build on the Monosemanticity Score (MonoScore) of \citet{pach2025sparse} to
quantify how consistently a latent responds to a single visual concept: a latent
is monosemantic if the images that strongly activate it are semantically
similar, and polysemantic if they are diverse and unrelated. The semantic
similarity space is defined by a frozen image encoder $E(\cdot)$; we use the same
encoder that produces the embeddings on which the SAE is trained (CLIP, SigLIP2,
or the supervised ViT-L/16, respectively), so that similarity is measured in the
same representation space the SAE decomposes. Concretely, $E$ yields
$\ell_2$-normalized embeddings $h^{(i)} = E(x^{(i)})/\lVert E(x^{(i)})\rVert_2
\in \mathbb{R}^{d}$, with cosine similarities
$s_{ii'} = {h^{(i)}}^{\top} h^{(i')} \in [-1,1]$ for all unordered pairs
$(i,i')$, $i<i'$. For comparability across latents, MonoScore first min--max
normalizes each latent's activations to its dynamic range,
\begin{equation}
\tilde{z}^{(i)}_j = \bigl(z^{(i)}_j - \min_{i'} z^{(i')}_j\bigr) \big/
\bigl(\max_{i'} z^{(i')}_j - \min_{i'} z^{(i')}_j\bigr) \in [0,1],
\end{equation}
and defines the score for latent $j$ as the activation-weighted average of
pairwise similarities,
\begin{equation}
\mathrm{MS}_j =
\frac{1}{\sum_{i<i'} \tilde{z}^{(i)}_j\,\tilde{z}^{(i')}_j}
\sum_{i<i'} \tilde{z}^{(i)}_j\,\tilde{z}^{(i')}_j\, s_{ii'}.
\end{equation}
We summarize the per-latent scores across latents by their mean
($\mathrm{M}_\mu$) and median ($\mathrm{M}_m$); higher is better. 

\paragraph{Class purity.}
As a label-grounded complement to MonoScore, class purity measures whether a
latent is associated with a dominant class among the samples that activate it.
The \emph{dominant class} of latent $j$ is the class most frequently represented
among its activating samples,
\[
c^{\star}_j = \operatorname*{arg\,max}_{c \in \{1,\dots,C\}}
\sum_{i \in A_j} \mathbf{1}\big[\,y^{(i)} = c\,\big].
\]
\emph{Binary purity} is the fraction of activating samples belonging to this
dominant class, counting every activation equally,
\[
\mathrm{Purity}^{\mathrm{bin}}_j =
\frac{1}{|A_j|} \sum_{i \in A_j} \mathbf{1}\big[\,y^{(i)} = c^{\star}_j\,\big].
\]
\emph{Weighted purity} uses the same dominant class $c^{\star}_j$ but weights
each activating sample by its activation strength, so that a latent's strongest
responses contribute more,
\[
\mathrm{Purity}^{\mathrm{wgt}}_j =
\frac{\sum_{i \in A_j}\, z^{(i)}_j\, \mathbf{1}\big[\,y^{(i)} = c^{\star}_j\,\big]}
{\sum_{i \in A_j}\, z^{(i)}_j } .
\]
Both are averaged over the active latents to give the reported scores,
\[
\mathrm{Purity}^{\bullet} = \frac{1}{|\mathcal{J}_{\mathrm{act}}|}
\sum_{j \in \mathcal{J}_{\mathrm{act}}} \mathrm{Purity}^{\bullet}_j,
\qquad \bullet \in \{\mathrm{bin}, \mathrm{wgt}\}.
\]
Higher is better for both. Both metrics use the \emph{same} dominant class
$c^{\star}_j$, determined by sample count; weighted purity differs only in that
its numerator and denominator are activation-mass-weighted rather than sample
counts.

\section{Hyperparameters}
\label{app:hyperparams}

\paragraph{General.}
Every model is trained for $50$ epochs with batch size $2048$ using Adam
(learning rate $10^{-4}$, $\epsilon = 6.25\times10^{-10}$) and gradient-norm
clipping at $1$. When a sparsity regularizer is applied, it is introduced only
after a $10$-epoch warmup, so that the Top-$k$ autoencoder first settles into a
useful reconstruction before the additional penalty takes effect. The
auxiliary-loss coefficient is fixed at $\alpha = 1/32$ throughout, and the
latent dictionary has $m = 8192$ units. Each configuration is trained with three
random seeds ($42, 43, 44$) to assess statistical variability. Training is
performed on a single NVIDIA H100 GPU.

\paragraph{Choice of the regularizer coefficient $\lambda$.}
For each (dataset, encoder, $k$, regularizer) configuration, we select
$\lambda$ by a search on the validation set: we retain the value giving the
highest mean monosemanticity while keeping the reconstruction error no worse
than the unregularized baseline. Results are reported on the test set, except
for label-dependent metrics on ImageNet-1K, whose test labels are not public;
for those we report on the validation set. Although the validation set is then
used both to select $\lambda$ and to compute label-dependent metrics such as
class purity, this does not constitute selection leakage: $\lambda$ is chosen
using only reconstruction ($R^2$) and the monosemanticity score, neither of
which uses labels, so the label-grounded metrics remain independent of the
selection criterion. The retained values of $\lambda$ are listed in
Table~\ref{tab:lambda}.

\begin{table}[t]
\centering
\small
\setlength{\tabcolsep}{5pt}
\begin{tabular}{lll cc}
\toprule
Encoder & Data & $k$ & Reg.~1 ($\ell_1$) & Reg.~2 ($\ell_1/\ell_2$) \\
\midrule
CLIP & ImageNet & 32 & 0.0003 & 0.001 \\
 &  & 64 & 0.0003 & 0.007 \\
 &  & 128 & 0.0001 & 0.003 \\
\addlinespace
 & Open Images & 32 & 0.0006 & 0.002 \\
 &  & 64 & 0.0008 & 0.004 \\
 &  & 128 & 0.001 & 0.005 \\
\midrule
SigLIP2 & ImageNet & 32 & 0.00008 & 0.0002 \\
 &  & 64 & 0.0001 & 0.00045 \\
 &  & 128 & 0.0001 & 0.0014 \\
\addlinespace
 & Open Images & 32 & 0.0001 & 0.002 \\
 &  & 64 & 0.0003 & 0.003 \\
 &  & 128 & 0.0006 & 0.001 \\
\midrule
ViT & ImageNet & 32 & 0.001 & 0.008 \\
 &  & 64 & 0.0001 & 0.007 \\
 &  & 128 & 0.0001 & 0.001 \\
\addlinespace
 & Open Images & 32 & 0.001 & 0.01 \\
 &  & 64 & 0.0006 & 0.02 \\
 &  & 128 & 0.0004 & 0.02 \\
\bottomrule
\end{tabular}
\caption{Selected regularizer coefficient $\lambda$ for each (encoder, dataset,
$k$) configuration, for Reg.~1 (off-support $\ell_1$) and Reg.~2 ($\ell_1/\ell_2$
ratio). Values correspond to the results reported in Table~1 of the
main paper.}
\label{tab:lambda}
\end{table}

\section{Datasets}
\label{app:datasets}

We evaluate on two large-scale image datasets that provide complementary
settings. ImageNet-1K \citep{russakovsky2015imagenet} is a single-label
classification benchmark with $1{,}000$ object categories, on which we compute
class-level semantic coherence and purity. Open Images V7 \citep{kuznetsova2020open} is a larger and more diverse collection whose
images span many object categories; we use it to test whether the
label-independent findings (reconstruction and monosemanticity) hold on a
broader distribution, and do not compute class purity on it. For both datasets
we operate on frozen image embeddings extracted from the three pretrained
encoders (CLIP ViT-L/14, SigLIP2, and the supervised ViT-L/16); the encoders
themselves are never updated. Sample counts per split are identical across the
three encoders and are given in Table~\ref{tab:splits}.

\begin{table}[t]
\centering
\small
\setlength{\tabcolsep}{8pt}
\begin{tabular}{lrr}
\toprule
Split & ImageNet-1K & Open Images V7 \\
\midrule
Train      & 1{,}281{,}167 & 507{,}444 \\
Validation & 50{,}000      & 41{,}691  \\
Test       & 100{,}000     & 126{,}020 \\
\bottomrule
\end{tabular}
\caption{Number of samples per split for the two datasets. Counts are identical
across all three feature extractors (CLIP, SigLIP2, ViT). ImageNet-1K uses the
canonical ILSVRC2012 splits; its official test split has no public ground-truth
labels, so label-dependent metrics on ImageNet-1K are reported on the validation
split. The Open Images V7 counts correspond to the ``Localized Narratives''
annotation subset.}
\label{tab:splits}
\end{table}

\section{Additional Results}
\label{app:additional}

This section reports results that complement the main paper: per-seed standard
deviations (Section~\ref{app:std}), zero-shot cross-dataset
transfer (Section~\ref{app:zeroshot}), inference-time $k$ robustness for the two
encoders omitted from the main text
(Section~\ref{app:robustness_other_encoders}), and additional qualitative
comparisons across monosemanticity ranks (Section~\ref{app:qualitative_ranks}).

\subsection{Standard Deviations for the Main Results}
\label{app:std}

Table~\ref{tab:std} reports the per-run standard deviations across the three
seeds $(42,43,44)$ for every entry of Table~1 in the main paper.

Table~\ref{tab:purity_std} likewise reports the per-run standard deviations
across the three seeds for the class-purity results (binary and weighted) of
Table~2 in the main paper.

\begin{table*}[t]
\centering
\small
\setlength{\tabcolsep}{3pt}
\begin{tabular}{lll cccc cccc cccc}
\toprule
& & & \multicolumn{4}{c}{$k=32$} & \multicolumn{4}{c}{$k=64$} & \multicolumn{4}{c}{$k=128$} \\
\cmidrule(lr){4-7} \cmidrule(lr){8-11} \cmidrule(lr){12-15}
Encoder & Data & Method & $R^2$ & M$_\mu$ & M$_m$ & Dead & $R^2$ & M$_\mu$ & M$_m$ & Dead & $R^2$ & M$_\mu$ & M$_m$ & Dead \\
\midrule
CLIP & ImageNet & Baseline & 0.001 & 0.000 & 0.000 & 1 & 0.000 & 0.001 & 0.001 & 1 & 0.000 & 0.000 & 0.001 & 1 \\
 &  & Reg.~1 & 0.001 & 0.001 & 0.001 & 5 & 0.000 & 0.000 & 0.001 & 1 & 0.000 & 0.001 & 0.001 & 1 \\
 &  & Reg.~2 & 0.000 & 0.000 & 0.000 & 8 & 0.000 & 0.000 & 0.000 & 4 & 0.000 & 0.000 & 0.001 & 1 \\
\addlinespace
 & Open Images & Baseline & 0.000 & 0.002 & 0.002 & 11 & 0.001 & 0.001 & 0.000 & 1 & 0.001 & 0.001 & 0.000 & 0 \\
 &  & Reg.~1 & 0.000 & 0.002 & 0.001 & 8 & 0.001 & 0.001 & 0.000 & 1 & 0.000 & 0.001 & 0.001 & 0 \\
 &  & Reg.~2 & 0.001 & 0.002 & 0.002 & 6 & 0.001 & 0.001 & 0.001 & 3 & 0.001 & 0.001 & 0.001 & 3 \\
\midrule
SigLIP2 & ImageNet & Baseline & 0.001 & 0.002 & 0.001 & 21 & 0.000 & 0.002 & 0.001 & 20 & 0.000 & 0.001 & 0.001 & 16 \\
 &  & Reg.~1 & 0.001 & 0.003 & 0.001 & 25 & 0.000 & 0.001 & 0.001 & 12 & 0.000 & 0.001 & 0.000 & 8 \\
 &  & Reg.~2 & 0.001 & 0.003 & 0.001 & 30 & 0.000 & 0.000 & 0.001 & 5 & 0.000 & 0.001 & 0.000 & 14 \\
\addlinespace
 & Open Images & Baseline & 0.001 & 0.002 & 0.002 & 13 & 0.001 & 0.001 & 0.001 & 12 & 0.001 & 0.000 & 0.001 & 6 \\
 &  & Reg.~1 & 0.000 & 0.002 & 0.001 & 5 & 0.001 & 0.001 & 0.001 & 6 & 0.000 & 0.000 & 0.001 & 2 \\
 &  & Reg.~2 & 0.000 & 0.002 & 0.002 & 34 & 0.001 & 0.003 & 0.001 & 42 & 0.001 & 0.001 & 0.000 & 9 \\
\midrule
ViT & ImageNet & Baseline & 0.003 & 0.001 & 0.002 & 0 & 0.001 & 0.000 & 0.001 & 0 & 0.001 & 0.000 & 0.001 & 0 \\
 &  & Reg.~1 & 0.002 & 0.001 & 0.001 & 1 & 0.001 & 0.001 & 0.000 & 0 & 0.001 & 0.001 & 0.001 & 0 \\
 &  & Reg.~2 & 0.000 & 0.001 & 0.001 & 0 & 0.000 & 0.001 & 0.003 & 0 & 0.001 & 0.000 & 0.001 & 0 \\
\addlinespace
 & Open Images & Baseline & 0.003 & 0.003 & 0.004 & 0 & 0.003 & 0.000 & 0.001 & 0 & 0.000 & 0.000 & 0.000 & 0 \\
 &  & Reg.~1 & 0.001 & 0.002 & 0.002 & 4 & 0.002 & 0.001 & 0.002 & 0 & 0.000 & 0.001 & 0.000 & 0 \\
 &  & Reg.~2 & 0.000 & 0.000 & 0.000 & 2 & 0.005 & 0.001 & 0.001 & 1 & 0.002 & 0.001 & 0.000 & 0 \\
\bottomrule
\end{tabular}
\caption{Per-run standard deviations across the three seeds ($42, 43, 44$) for
the results reported in Table~1 of the main paper: reconstruction
($R^2$), mean and median monosemanticity (M$_\mu$, M$_m$), and number of dead
neurons, for the Top-$k$ baseline and our two regularizers---Reg.~1 (off-support
$\ell_1$) and Reg.~2 ($\ell_1/\ell_2$ ratio).}
\label{tab:std}
\end{table*}

\begin{table}[tb]
\centering
\small
\setlength{\tabcolsep}{3pt}
\begin{tabular}{ll ccc ccc}
\toprule
& & \multicolumn{3}{c}{Binary purity} & \multicolumn{3}{c}{Weighted purity} \\
\cmidrule(lr){3-5} \cmidrule(lr){6-8}
Enc. & $k$ & Base & Reg.~1 & Reg.~2 & Base & Reg.~1 & Reg.~2 \\
\midrule
CLIP & 32 & 0.0039 & 0.0025 & 0.0034 & 0.0044 & 0.0023 & 0.0036 \\
 & 64 & 0.0008 & 0.0014 & 0.0011 & 0.0011 & 0.0013 & 0.0014 \\
 & 128 & 0.0002 & 0.0019 & 0.0001 & 0.0008 & 0.0023 & 0.0005 \\
\midrule
SigLIP2 & 32 & 0.0029 & 0.0032 & 0.0046 & 0.0022 & 0.0021 & 0.0029 \\
 & 64 & 0.0015 & 0.0032 & 0.0006 & 0.0009 & 0.0026 & 0.0004 \\
 & 128 & 0.0007 & 0.0014 & 0.0011 & 0.0005 & 0.0015 & 0.0007 \\
\midrule
ViT & 32 & 0.0007 & 0.0007 & 0.0040 & 0.0010 & 0.0022 & 0.0010 \\
 & 64 & 0.0009 & 0.0013 & 0.0023 & 0.0008 & 0.0010 & 0.0011 \\
 & 128 & 0.0003 & 0.0006 & 0.0004 & 0.0005 & 0.0013 & 0.0010 \\
\bottomrule
\end{tabular}
\caption{Per-run standard deviations across the three seeds ($42, 43, 44$) for
the class-purity results reported in Table~2 of the main paper: binary
and weighted purity on the ImageNet-1K validation set, for the Top-$k$ baseline
(Base) and our two regularizers---Reg.~1 (off-support $\ell_1$) and Reg.~2
($\ell_1/\ell_2$ ratio).}
\label{tab:purity_std}
\end{table}

\subsection{Zero-Shot Cross-Dataset Transfer}
\label{app:zeroshot}

Table~\ref{tab:zeroshot} reports the full zero-shot transfer results summarized
in the main text: SAEs trained on one dataset and evaluated on the other, in
both directions. $\lambda$ values selected for the regularized models are still the ones listed in Table~\ref{tab:lambda}.

\begin{table*}[t]
\centering
\small
\setlength{\tabcolsep}{3pt}
\begin{tabular}{lll cccc cccc cccc}
\toprule
& & & \multicolumn{4}{c}{$k=32$} & \multicolumn{4}{c}{$k=64$} & \multicolumn{4}{c}{$k=128$} \\
\cmidrule(lr){4-7} \cmidrule(lr){8-11} \cmidrule(lr){12-15}
Encoder & Transfer & Method & $R^2$ & M$_\mu$ & M$_m$ & Dead & $R^2$ & M$_\mu$ & M$_m$ & Dead & $R^2$ & M$_\mu$ & M$_m$ & Dead \\
\midrule
CLIP & IN$\to$OI & Baseline & 0.694 & 0.464 & 0.457 & 23 & 0.759 & 0.427 & 0.407 & 4 & 0.825 & 0.389 & 0.362 & 2 \\
 &  & Reg.~1 & \underline{0.695} & \textbf{0.484} & \textbf{0.482} & 26 & \underline{0.760} & \textbf{0.460} & \textbf{0.450} & 1 & \underline{0.827} & \textbf{0.404} & \textbf{0.382} & 1 \\
 &  & Reg.~2 & \textbf{0.698} & \underline{0.466} & \underline{0.462} & 52 & \textbf{0.762} & \underline{0.446} & \underline{0.431} & 12 & \textbf{0.829} & \underline{0.397} & \underline{0.371} & 2 \\
\addlinespace
 & OI$\to$IN & Baseline & 0.670 & 0.433 & 0.436 & 54 & 0.735 & 0.410 & 0.401 & 4 & 0.793 & 0.387 & 0.379 & 0 \\
 &  & Reg.~1 & \underline{0.671} & \textbf{0.448} & \textbf{0.463} & 132 & \underline{0.735} & \textbf{0.438} & \textbf{0.433} & 6 & \underline{0.798} & \textbf{0.407} & \textbf{0.396} & 0 \\
 &  & Reg.~2 & \textbf{0.677} & \underline{0.437} & \underline{0.445} & 104 & \textbf{0.744} & \underline{0.420} & \underline{0.412} & 11 & \textbf{0.803} & \underline{0.392} & \underline{0.384} & 4 \\
\midrule
SigLIP2 & IN$\to$OI & Baseline & \underline{0.668} & \underline{0.564} & \underline{0.574} & 195 & \underline{0.742} & \underline{0.552} & 0.554 & 199 & \underline{0.805} & 0.533 & 0.525 & 118 \\
 &  & Reg.~1 & 0.668 & \textbf{0.568} & \textbf{0.580} & 209 & 0.742 & \textbf{0.563} & \textbf{0.567} & 163 & 0.804 & \textbf{0.550} & \textbf{0.543} & 72 \\
 &  & Reg.~2 & \textbf{0.670} & 0.562 & 0.574 & 212 & \textbf{0.746} & 0.550 & \underline{0.555} & 239 & \textbf{0.811} & \underline{0.535} & \underline{0.529} & 158 \\
\addlinespace
 & OI$\to$IN & Baseline & 0.628 & \underline{0.530} & 0.556 & 256 & 0.705 & \underline{0.525} & 0.529 & 120 & 0.764 & \underline{0.510} & 0.506 & 48 \\
 &  & Reg.~1 & \underline{0.631} & \textbf{0.531} & \textbf{0.564} & 303 & \underline{0.706} & \textbf{0.542} & \textbf{0.547} & 78 & \underline{0.765} & \textbf{0.528} & \textbf{0.522} & 8 \\
 &  & Reg.~2 & \textbf{0.636} & 0.525 & \underline{0.558} & 360 & \textbf{0.714} & 0.522 & \underline{0.535} & 259 & \textbf{0.770} & 0.509 & \underline{0.507} & 86 \\
\midrule
ViT & IN$\to$OI & Baseline & \underline{0.632} & 0.261 & 0.241 & 1 & \underline{0.686} & 0.169 & 0.147 & 0 & 0.735 & 0.109 & 0.088 & 0 \\
 &  & Reg.~1 & 0.629 & \textbf{0.360} & \textbf{0.343} & 0 & 0.686 & \textbf{0.218} & \textbf{0.197} & 0 & \underline{0.736} & \textbf{0.132} & \textbf{0.112} & 0 \\
 &  & Reg.~2 & \textbf{0.674} & \underline{0.322} & \underline{0.301} & 3 & \textbf{0.714} & \underline{0.217} & \underline{0.193} & 1 & \textbf{0.744} & \underline{0.116} & \underline{0.094} & 0 \\
\addlinespace
 & OI$\to$IN & Baseline & 0.678 & 0.184 & 0.157 & 0 & 0.708 & 0.128 & 0.102 & 0 & 0.733 & 0.092 & 0.073 & 0 \\
 &  & Reg.~1 & \underline{0.682} & \textbf{0.291} & \textbf{0.260} & 8 & \underline{0.714} & \textbf{0.211} & \textbf{0.180} & 0 & \underline{0.739} & \textbf{0.144} & \textbf{0.119} & 0 \\
 &  & Reg.~2 & \textbf{0.706} & \underline{0.261} & \underline{0.230} & 9 & \textbf{0.734} & \underline{0.187} & \underline{0.156} & 1 & \textbf{0.754} & \underline{0.113} & \underline{0.090} & 0 \\
\bottomrule
\end{tabular}
\caption{Zero-shot cross-dataset transfer. Reconstruction ($R^2$), mean and
median monosemanticity (M$_\mu$, M$_m$), and number of dead neurons for the
Top-$k$ baseline and our two regularizers, Reg.~1 (off-support $\ell_1$) and
Reg.~2 ($\ell_1/\ell_2$ ratio), when SAEs trained on one dataset are evaluated
on the other. IN$\to$OI denotes training on ImageNet-1K and evaluating on Open
Images V7, and OI$\to$IN the reverse. All values are means over three seeds
($42,43,44$).}
\label{tab:zeroshot}
\end{table*}

\subsection{Robustness to the Inference-Time $k$: Other Encoders}
\label{app:robustness_other_encoders}

\begin{figure}[tb]
    \centering
    \begin{subfigure}{\columnwidth}
        \centering
        \includegraphics[width=\columnwidth]{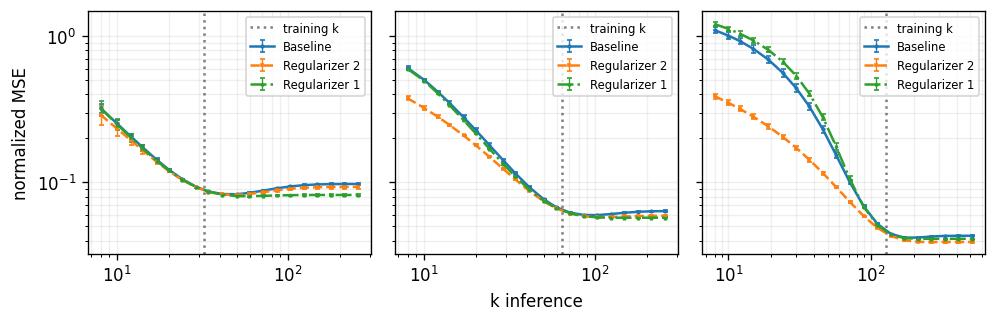}
        \caption{SigLIP2}
        \label{fig:robustness_siglip}
    \end{subfigure}

    \vspace{0.5em}

    \begin{subfigure}{\columnwidth}
        \centering
        \includegraphics[width=\columnwidth]{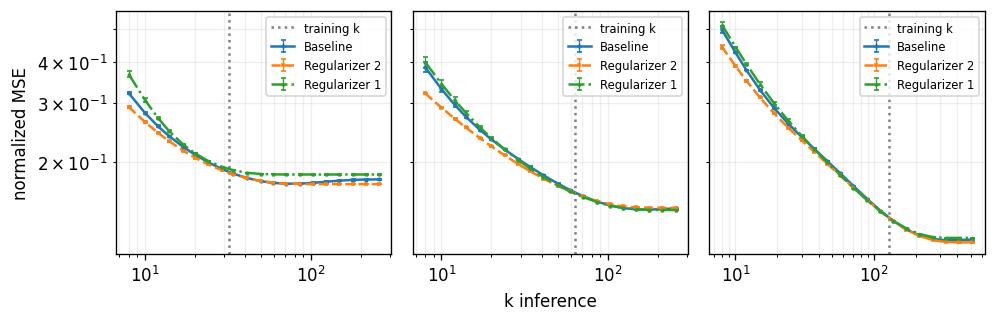}
        \caption{Supervised ViT-L/16}
        \label{fig:robustness_vit}
    \end{subfigure}

    \caption{Robustness to the inference-time $k$ on ImageNet-1K for the two
    remaining encoders, following the protocol of Figure~7 of the main
    paper (which shows CLIP ViT-L/14). (a)~SigLIP2; (b)~the supervised
    ViT-L/16.}
    \label{fig:robustness_other_encoders}
\end{figure}

Figures~\ref{fig:robustness_siglip} and~\ref{fig:robustness_vit} report the
inference-time $k$ robustness results of Figure~7 of the main paper for
the two remaining encoders, SigLIP2 and the supervised ViT-L/16.

\subsection{Additional Qualitative Comparisons}
\label{app:qualitative_ranks}

\begin{figure*}[t]
    \centering
    \begin{subfigure}{0.49\textwidth}
        \centering
        \includegraphics[width=\textwidth]{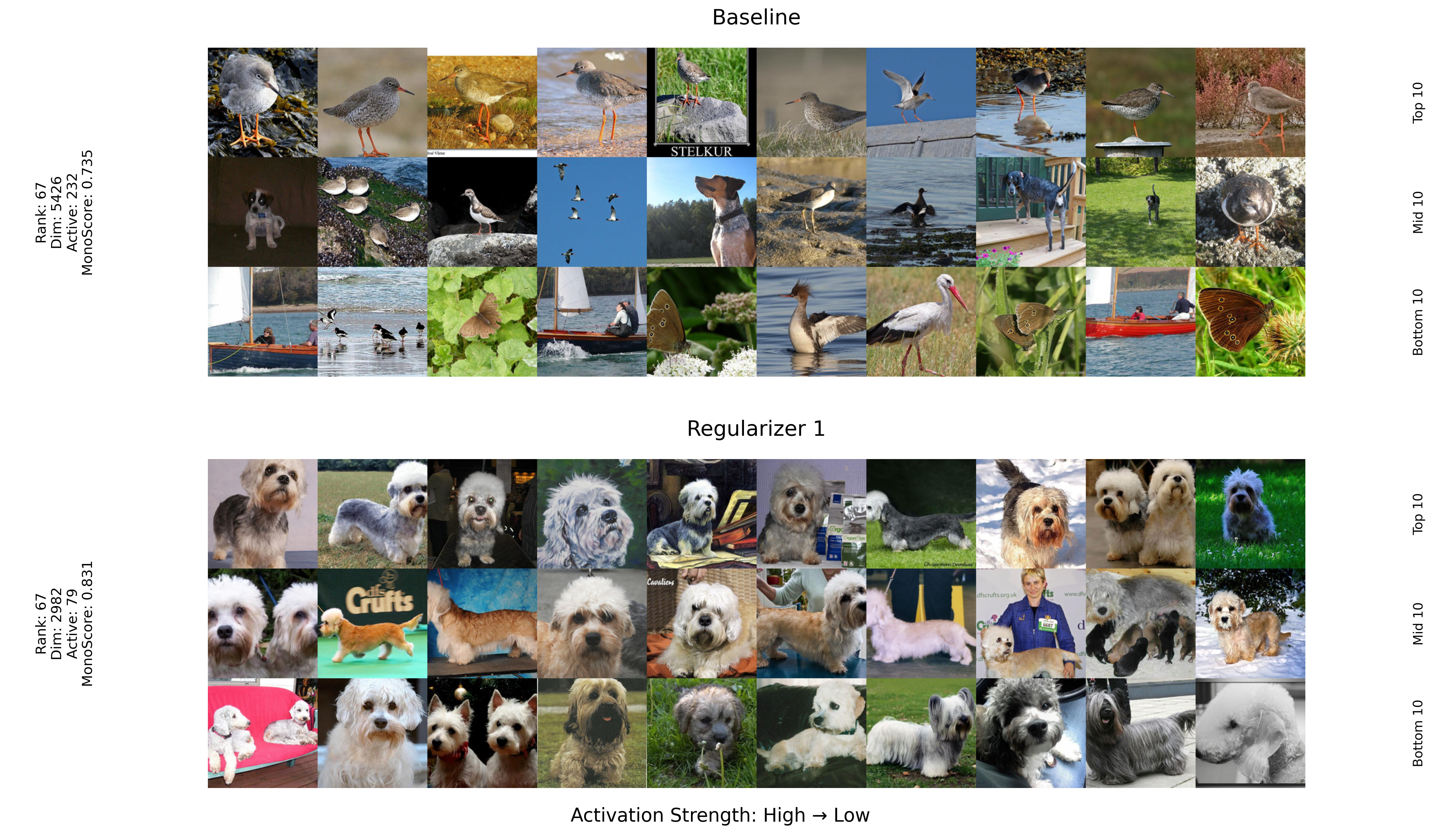}
        \caption{Rank 67}
        \label{fig:qual_rank_67}
    \end{subfigure}\hfill
    \begin{subfigure}{0.49\textwidth}
        \centering
        \includegraphics[width=\textwidth]{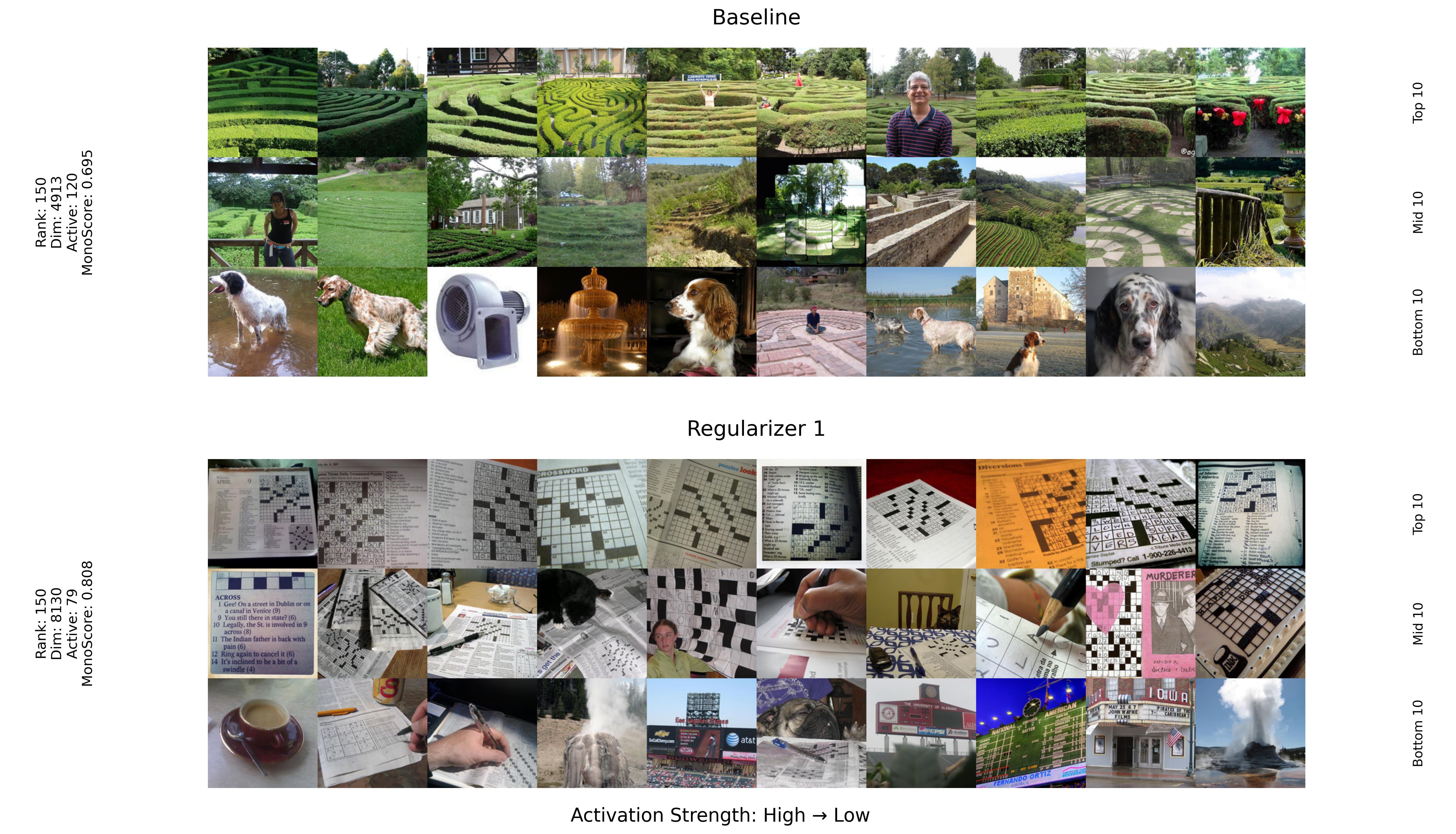}
        \caption{Rank 150}
        \label{fig:qual_rank_150}
    \end{subfigure}
    \caption{Qualitative comparison at matched monosemanticity ranks 67 and 150 (ViT-L/16, $k=32$). In each panel, the top block is a baseline latent and the bottom block the Regularizer~1 latent at the same rank.}
    \label{fig:qual_67_150}
\end{figure*}

\begin{figure*}[t]
    \centering
    \begin{subfigure}{0.49\textwidth}
        \centering
        \includegraphics[width=\textwidth]{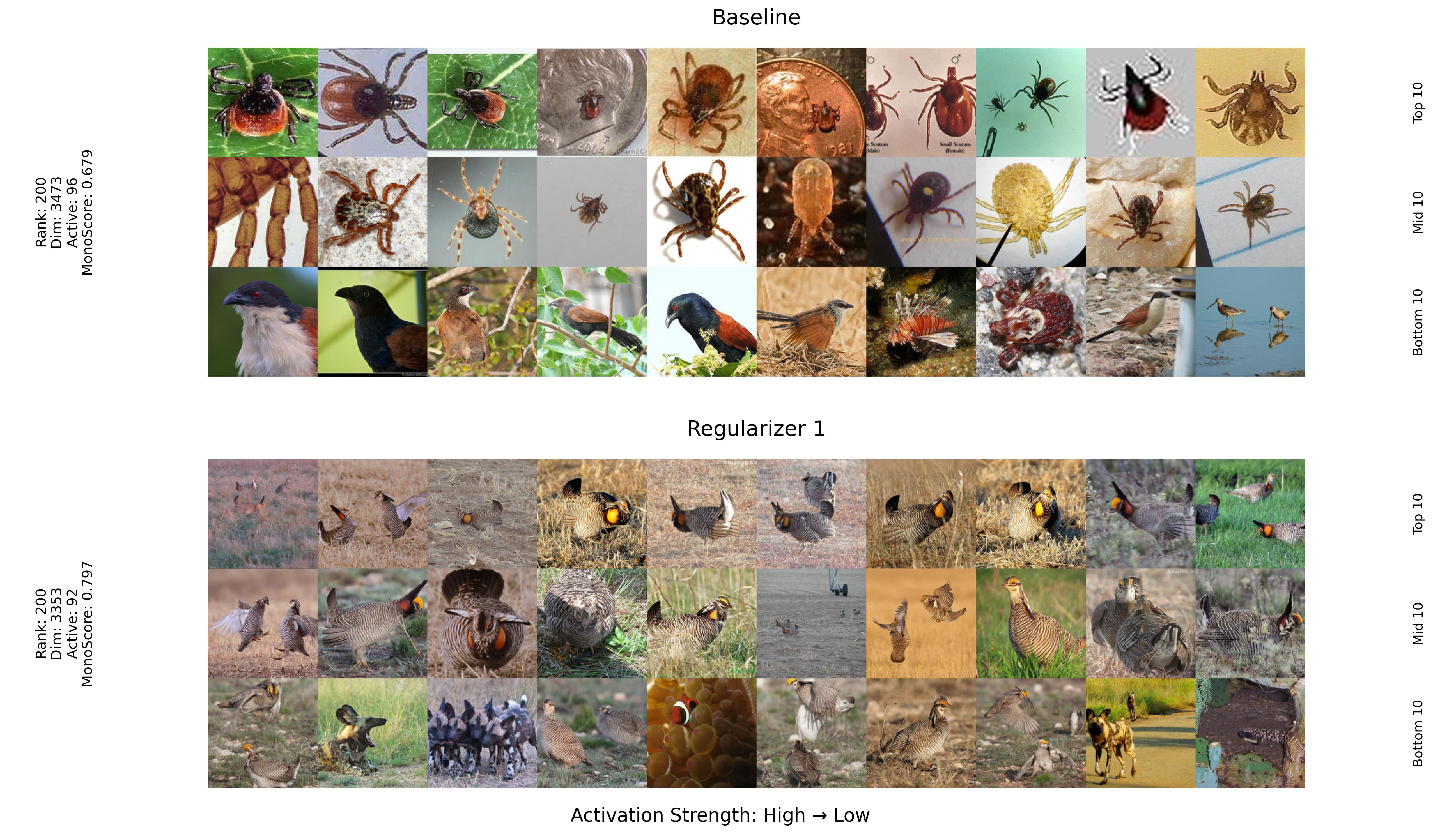}
        \caption{Rank 200}
        \label{fig:qual_rank_200}
    \end{subfigure}\hfill
    \begin{subfigure}{0.49\textwidth}
        \centering
        \includegraphics[width=\textwidth]{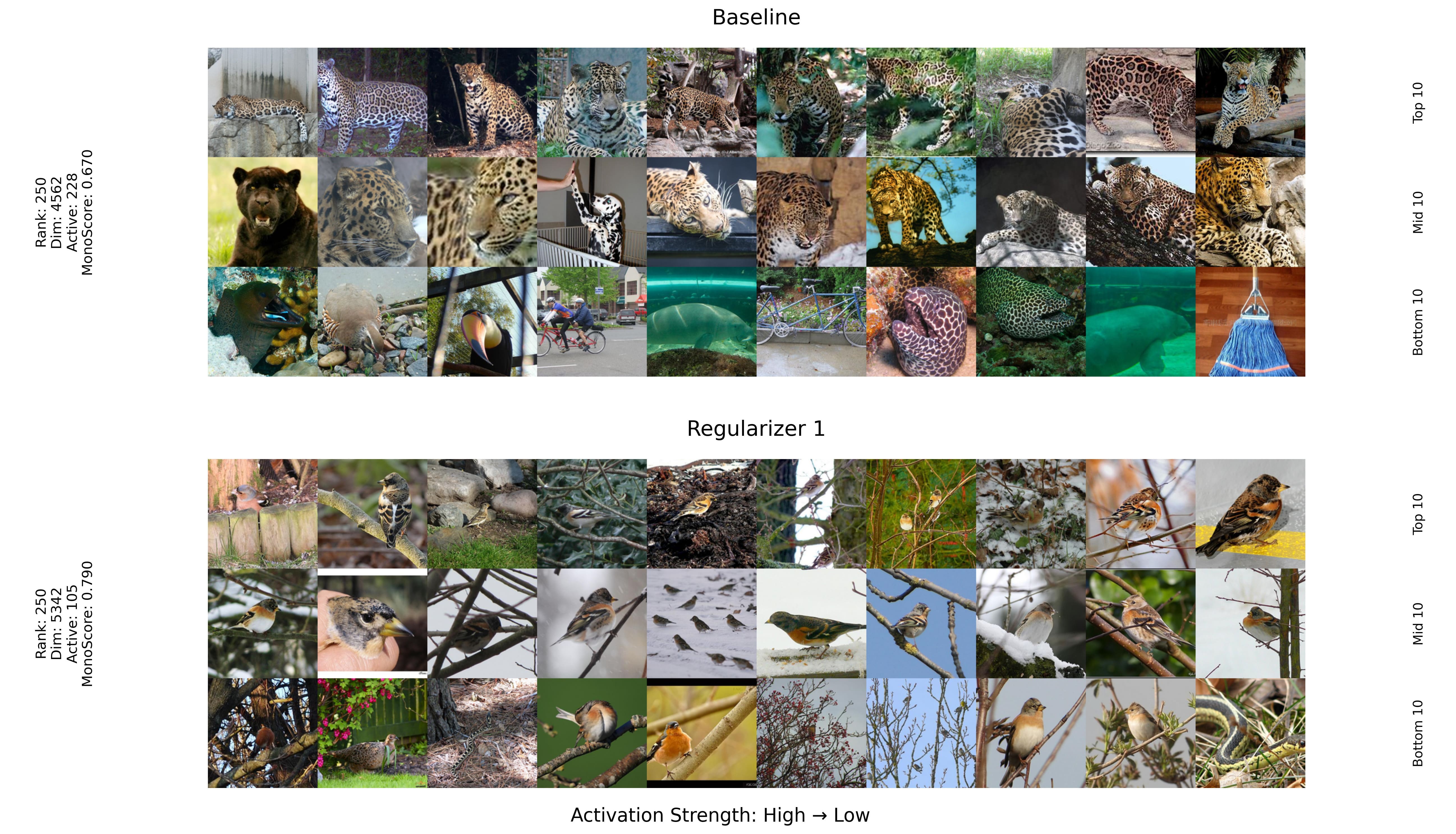}
        \caption{Rank 250}
        \label{fig:qual_rank_250}
    \end{subfigure}
    \caption{Qualitative comparison at matched monosemanticity ranks 200 and 250 (ViT-L/16, $k=32$). In each panel, the top block is a baseline latent and the bottom block the Regularizer~1 latent at the same rank.}
    \label{fig:qual_200_250}
\end{figure*}

\begin{figure*}[t]
    \centering
    \begin{subfigure}{0.49\textwidth}
        \centering
        \includegraphics[width=\textwidth]{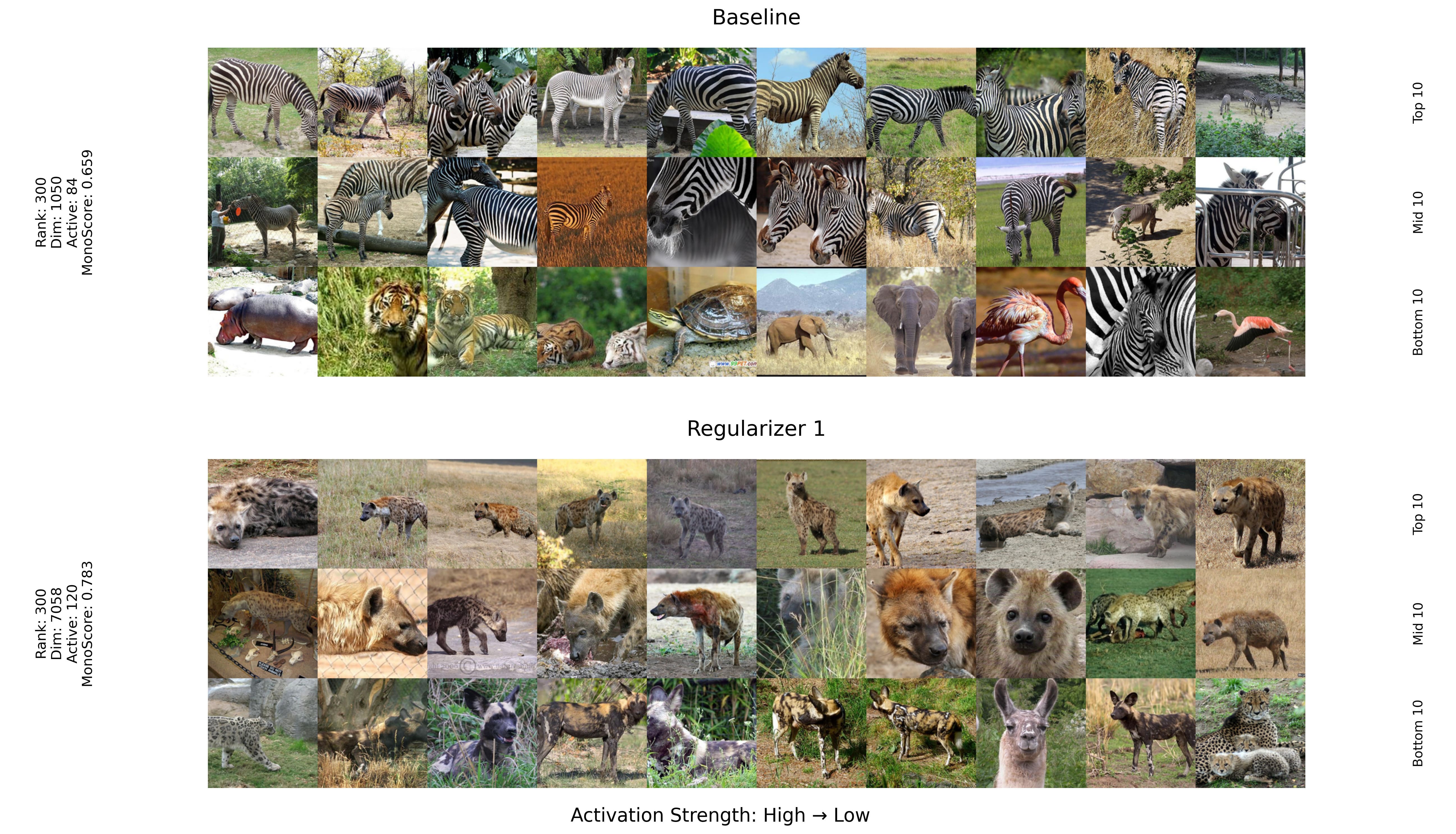}
        \caption{Rank 300}
        \label{fig:qual_rank_300}
    \end{subfigure}\hfill
    \begin{subfigure}{0.49\textwidth}
        \centering
        \includegraphics[width=\textwidth]{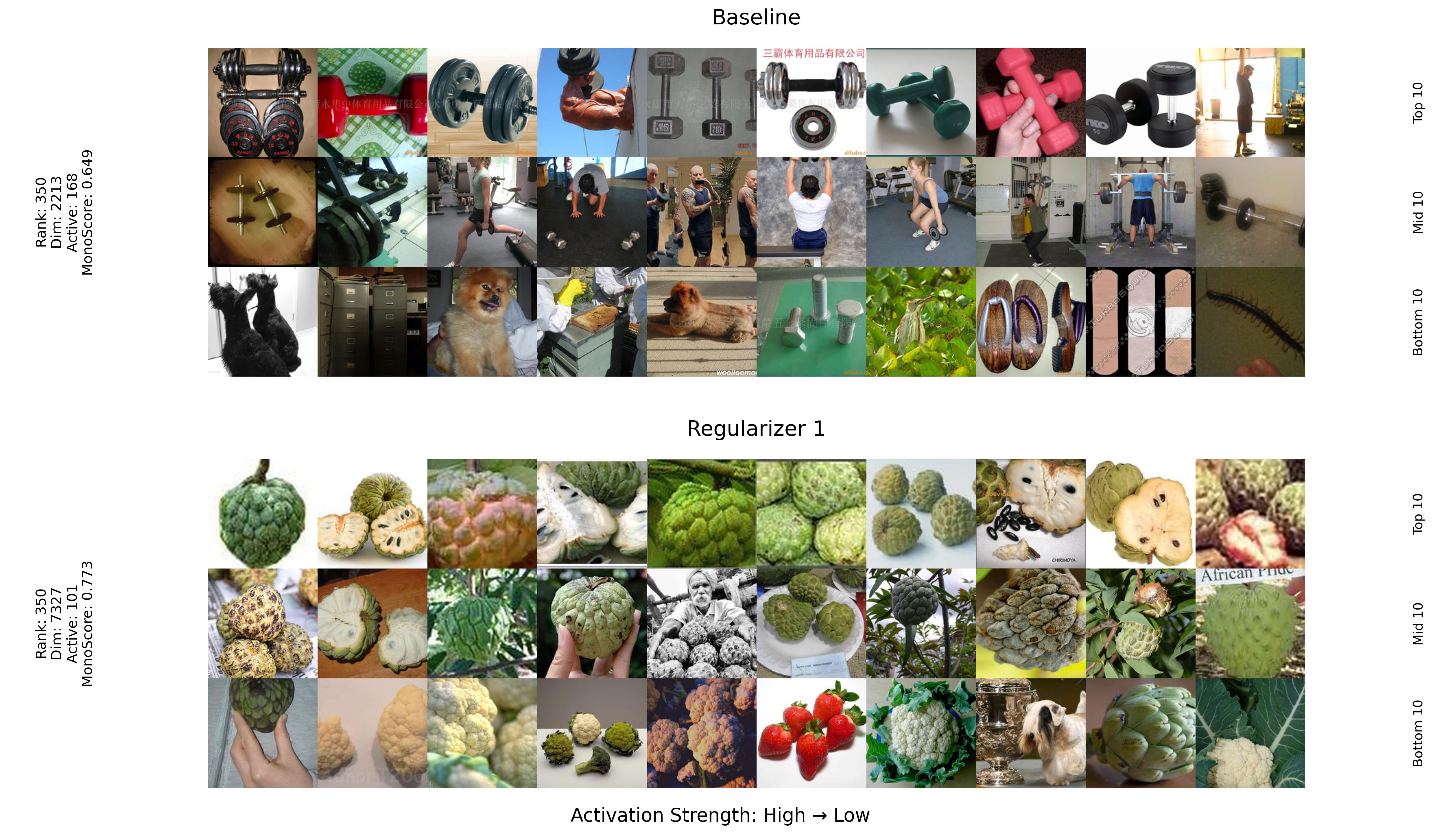}
        \caption{Rank 350}
        \label{fig:qual_rank_350}
    \end{subfigure}
    \caption{Qualitative comparison at matched monosemanticity ranks 300 and 350 (ViT-L/16, $k=32$). In each panel, the top block is a baseline latent and the bottom block the Regularizer~1 latent at the same rank.}
    \label{fig:qual_300_350}
\end{figure*}

Figures~\ref{fig:qual_rank_67}--\ref{fig:qual_rank_350} extend the rank-matched
qualitative comparison of the main text (Figure~4) to a range of
monosemanticity ranks. In each figure, the top block shows a baseline latent and
the bottom block the Regularizer~1 (off-support $\ell_1$) latent at the same
test-set monosemanticity rank; rows show the Top-10, Mid-10, and Bottom-10
activating images, ordered by decreasing activation strength. We select ranks in
two ways. Rank~67 (Figure~\ref{fig:qual_rank_67}) is chosen as an illustrative
best case: the regularized latent at this rank is perfectly semantically consistent across
its entire activation range, with the same human-interpretable concept present
from the Top-10 down to the Bottom-10 activating images. The remaining ranks,
$150$ through $350$ in steps of $50$
(Figures~\ref{fig:qual_rank_150}--\ref{fig:qual_rank_350}), are instead a fixed,
predetermined grid chosen without inspecting the latents beforehand; showing this
systematic sweep rather than hand-picked examples guards against selection bias
and demonstrates that the improvement in coherence is not confined to the most
visually favorable units.

\end{document}